%% file: custom.tex
\UseRawInputEncoding
\pdfoutput=1

\documentclass[11pt]{article}

\usepackage[final]{acl}

\usepackage{times}
\usepackage{latexsym}

\usepackage[T1]{fontenc}

\usepackage[utf8]{inputenc}

\usepackage{microtype}

\usepackage{inconsolata}

\usepackage{graphicx}
\usepackage{booktabs} 
\usepackage{colortbl}
\usepackage{caption}
\usepackage{hyperref}
\usepackage{algorithm}
\usepackage{algorithmicx}  
\usepackage{algcompatible} 
\usepackage{amsmath}
\usepackage{mathtools} 
\usepackage{float}  
\usepackage{lipsum} 
\usepackage{listings}
\usepackage{subcaption}
\usepackage{amssymb} 
\usepackage{enumitem}
\usepackage{makecell} 
\usepackage{multirow}
\usepackage{array} 
\usepackage{tabularx}

\title{PASS-FC: Progressive and Adaptive Search Scheme for Fact Checking of Comprehensive Claims}

\author{Ziyu Zhuang \\
  Trip.com Group  \\
  \texttt{royzhuang1124@gmail.com} 
  }

\begin{document}
\maketitle

\begin{abstract}
Automated fact-checking (AFC) still falters on claims that are time-sensitive, entity-ambiguous, or buried beneath noisy search-engine results. We present \textbf{PASS-FC}, a Progressive and Adaptive Search Scheme for Fact Checking. Each atomic claim is first \emph{grounded} with a precise time span and disambiguated entity descriptors. An adaptive search loop then issues structured queries, filters domains through credible-source selection, and expands queries cross-lingually; when necessary, a lightweight reflection routine restarts the loop. Experiments on six benchmarks—covering general knowledge, scientific literature, real-world events, and ten languages—show that PASS-FC consistently outperforms prior systems, even those powered by larger backbone LLMs. On the multilingual X-FACT set, performance of different languages partially correlates with typological closeness to English, and forcing the model to reason in low-resource languages degrades accuracy. Ablations highlight the importance of temporal grounding and the adaptive search scheme, while detailed analysis shows that cross-lingual retrieval contributes genuinely new evidence. Code and full results will be released to facilitate further research.
\end{abstract}

\section{Introduction}

\input{section/introduction}

\section{Related Work}

\input{section/related_work}

\section{PASS-FC}

\input{section/method}

\section{Experiments}

\input{section/exp}

\section*{Limitations}
While it provides a reasonable definition of atomic facts, the framework only checks factual claims within the processed text. It is unable to detect truthful but irrelevant responses or evaluate model refusals to answer, limiting its scope in assessing overall response quality.


Preliminary observations suggest a possible antagonistic relationship between claim decomposition and iterative verification. This was noted in experiments where the performance gain from iterative verification was less pronounced on datasets like FactBench, which already incorporate claim decomposition. Further experiments are needed to verify this hypothesis and understand the interplay between these components.



\bibliography{custom}

\input{section/Appendix}

\end{document}

%% file: section/introduction.tex
\begin{figure}[t]
  \includegraphics[width=\columnwidth]{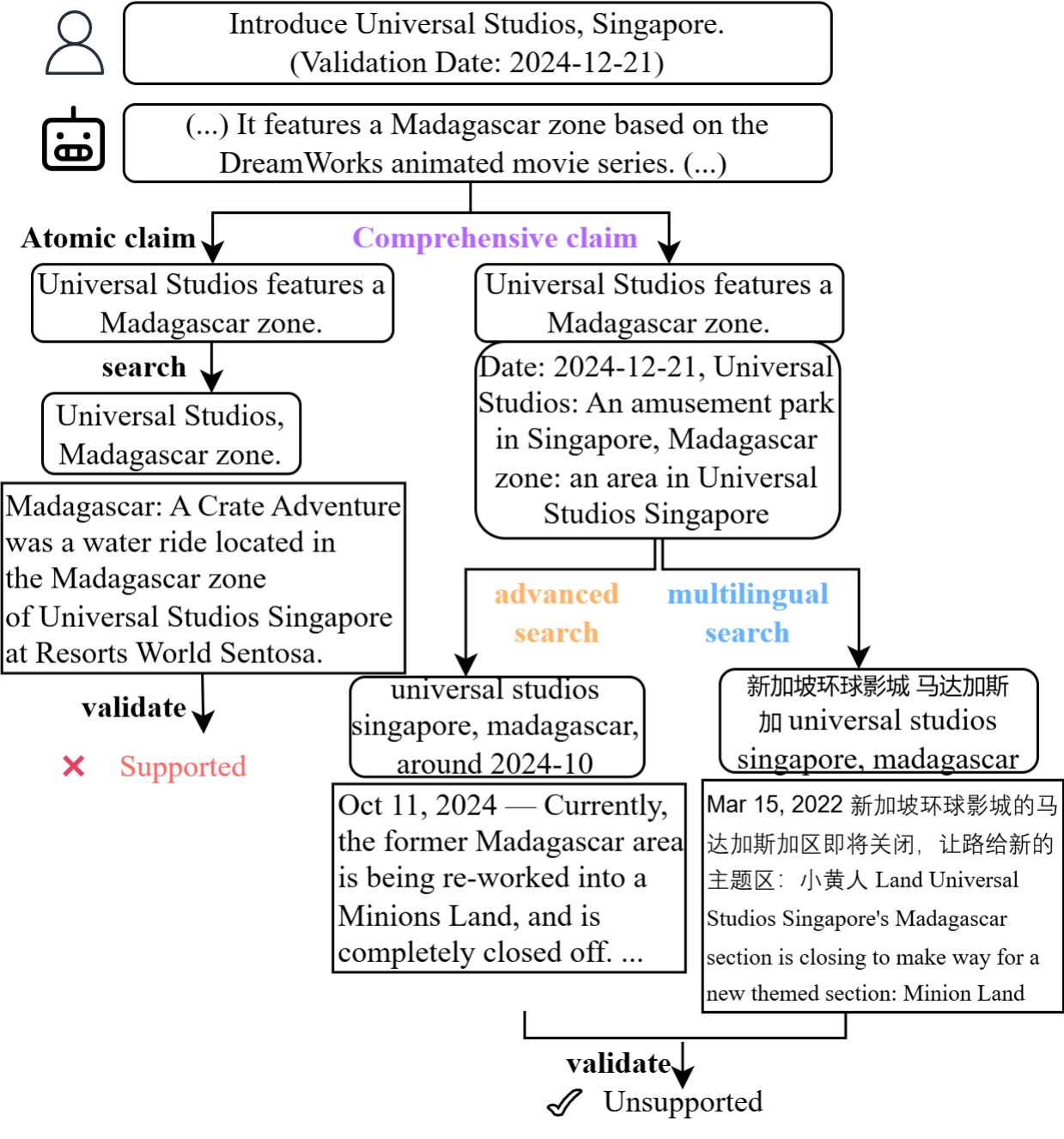}
  \captionsetup{font=small}
  \caption{Workflow comparison between a traditional fact-checking pipeline (left) and PASS-FC (right). PASS-FC enriches atomic claims with temporal and entity information, then employs advanced and multilingual search to collect relevant and sufficient evidence. Text inside rectangles shows retrieved snippets.}
  \label{fig:intro_show}
\end{figure}

The Web and large language models (LLMs) have created an information deluge, but also a minefield of inaccuracies \citep{webretrieve,factcheckgpt}. This surge in information has brought the critical task of fact-checking to the forefront \citep{concrete}. Standard approaches for automated fact-checking comprise three stages \citep{factool, safe, questgen}: (1) decomposing text into \emph{atomic claims}\footnote{An atomic claim is a short sentence conveying a single piece of information, as defined by Factscore \citep{factscore}.}; (2) searching the Web for relevant evidence; and (3) verifying each claim against that evidence. Unfortunately, two persistent bottlenecks still limit their reliability (Figure~\ref{fig:intro_show}).

\paragraph{Bottleneck1: missing context.}
The canonical atomic claim omits \emph{when} a statement is intended to hold and \emph{which} real-world entity it references \citep{molecularf,entitylink}. The left half of Figure~\ref{fig:intro_show} illustrates the pitfall: "Universal Studios features a Madagascar Zone." was once true but false in 2024, yet a date-free checker may still retrieve outdated corroboration \citep{molecularf,entitylink}.


\paragraph{Bottleneck~2: brittle search.}
Question-generation modules in AFC must operate under uncertain retrieval outcomes \citep{tot}. Vast, heterogeneous Web indices make it unlikely that single, naïve queries will surface relevant, credible, and sufficient evidence \citep{autorag,veriscore}. Information-seeking scenarios often require multiple iterations \citep{baleen,knowhalu}, advanced operators, domain filters, or multilingual queries before trustworthy sources emerge \citep{claimdecomp,varifocal,averitec} (as in the right path of Figure~\ref{fig:intro_show}). Existing systems \citep{safe,selfchecker} typically rely on simple query generation and hope that subsequent reflection alone can instruct it to uncover a workable verification path.


We address these challenges with \textbf{PASS-FC}. Each atomic fact is augmented with a temporal span and unique entity descriptors (§\ref{sec:claimaug}). Our \emph{adaptive search scheme} (§\ref{sec:querygen}) then crafts structured queries, selects credible domains, and, when needed, expands the search into other languages, progressively shrinking the evidence gap. After every round a reflection module (§\ref{sec:reflect}) inspects the verdict and either stops or issues a new tool plan. The system finally outputs a veracity label plus a concise rationale.

We first tune hyperparameters on a separate development set, revealing insights such as the optimal number of evidence passages and the varying benefits of reflection for different base models (Figure~\ref{fig:exp_2}c). With these settings, PASS-FC exceeds all baselines on six benchmarks, even outperforming systems that use stronger backbone LLMs (\S\ref{exp:exp1}). On the multilingual X-FACT set, performance of different languages partially correlates with typological distance from English, and forcing the model to reason in low-resource languages reduces accuracy (Table~\ref{tab:xfact_res}). Ablations quantify the impact of each module, underscoring the importance of temporal grounding and the adaptive search scheme (\S\ref{exp:exp3}). Further analysis shows that foreign-language evidence adds genuinely new support (\S\ref{exp:xle_analysis} and \S\ref{exp:case_analysis}).


\paragraph{Contributions}
\begin{itemize}[leftmargin=*]
    \item We propose PASS-FC, the first framework that unifies temporal/entity grounding with an adaptive, multilingual search loop for fact-checking.
    \item Extensive experiments demonstrate state-of-the-art accuracy on general, scientific, real-world, and multilingual benchmarks, even when PASS-FC uses smaller backbone LLMs.
    \item We release code, prompts, and complete results, and provide practical insights into evidence quantity, reflection triggers, multilingual fact-checking, and the role of cross-lingual evidence.
\end{itemize}

%% file: section/related_work.tex
\begin{table*}[ht]
\scriptsize  
\centering
\resizebox{\textwidth}{!}{%
\begin{tabular}{l|>{\centering\arraybackslash}p{2cm}>{\centering\arraybackslash}p{2cm}>{\centering\arraybackslash}p{2cm}>{\centering\arraybackslash}p{2cm}>{\centering\arraybackslash}p{2cm}}
\toprule
\textbf{Models} & \textbf{Structured Query Generation} & \textbf{Credible-Source Selection} & \textbf{Cross-Lingual Expansion} & \textbf{Temporal Grounding} & \textbf{Reflection} \\
\midrule
FacTool \citep{factool}& - & - & - & - & - \\
KnowHalu \citep{knowhalu}& - & - & - & - & $\checkmark$ \\
FOLK \citep{folk}& - & search in wikipedia & - & - & - \\
FactScore \citep{factscore}& - & - & - & - & - \\
ProgramFC \citep{programfc}& - & - & - & - & $\checkmark$ \\
Self-Checker \citep{selfchecker}& - & - & - & - & $\checkmark$ \\
SAFE \citep{safe}& - & - & - & - & $\checkmark$ \\
VERISCORE \citep{veriscore}& - & - & - & - & $\checkmark$ \\
Hiss \citep{hiss}& - & - & - & - & $\checkmark$ \\
PACAR \citep{zhao-etal-2024-pacar-automated}& - & - & - & - & $\checkmark$ \\
Rafts \citep{rafts}& - & - & - & - & - \\
MiniCheck \citep{minicheck}& - & - & - & - & - \\
FactCheckGPT \citep{factcheckgpt}& - & - & - & - & $\checkmark$ \\
DEFAME \citep{defame}& - & - & - & - & $\checkmark$ \\
FIRE \citep{fire}& - & - & - & - & $\checkmark$ \\
\midrule
PASS-FC (Our Method) & $\checkmark$ & $\checkmark$ & $\checkmark$ & $\checkmark$ & $\checkmark$ \\
\bottomrule
\end{tabular}
}
\captionsetup{font=small}
\caption{Comparison of fact-check frameworks across various capabilities. We claim all of them except reflection as our initial contribution to the automatic fact-checking (AFC) task.}
\label{tab:model_comparison}
\end{table*}

\noindent \textbf{Atomic Claim and its Grounding}
Effective fact verification requires a clear definition of facts \citep{afacta}, often necessitating the identification of atomic claims within longer texts. Numerous studies \citep{decomdilemma, fleek, factscore, veriscore} have proposed definitions for atomic facts. \citet{entitylink} first highlighted the issue with atomic facts in which excessive atomization can lead to entity ambiguity. They mitigate this with entity linking. \citet{molecularf} introduced the concept of molecular facts, refining the approach by expanding entity and event descriptions within atomic facts. While these methods have shown promise, they have primarily been tested in biographical generation tasks. Our work distinguishes itself in two ways: (1) Beyond supplementing entity groundings, we also consider \emph{temporal grounding}: an explicit date or time span that anchors truth to the correct period, preventing evidence drift. (2) We extend beyond previous studies by evaluating the impact of these enhancements on end-to-end fact verification in real-world scenarios. Table~\ref{tab:model_comparison} shows that PASS-FC is the first system that brings temporal grounding into the AFC pipeline.

\noindent \textbf{Query Generation for Fact Verification}
To address real-world information demands \citep{mulan, realtimeqa}, most frameworks \citep{rarr, wildevd, markovvalidate} generate queries to retrieve relevant knowledge from search engines like Google. Typical methods \citep{factool, veriscore, factcheckgpt} rely on few-shot examples to prompt LLMs to ask direct or entity attributes-based questions \citep{knowhalu, folk}. Yet human studies \citep{averitec} reveal large variance in good queries, with an average similarity of 0.25 between effective retrieval strategies across different individuals. Learning that diversity helps \citep{varifocal,questgen}, but existing datasets \citep{qabrief,averitec,claimdecomp,faviq} cover only narrow domains. PASS-FC enforces three constraints before any query is sent: (i) \emph{Credible-Source Selection} keeps only trustworthy domains, (ii) \emph{Cross-Lingual Expansion} adds languages likely to contain fresh evidence, and (iii) \emph{Structured Query Generation} composes Boolean queries that bind the claim, time, sources, and languages together. This adaptive search scheme is unique among current AFC systems (Table~\ref{tab:model_comparison}).


%% file: section/method.tex
\definecolor{lightblue}{rgb}{1,0.5,0}

\begin{figure*}[t]
  \centering
  \includegraphics[scale=0.5]{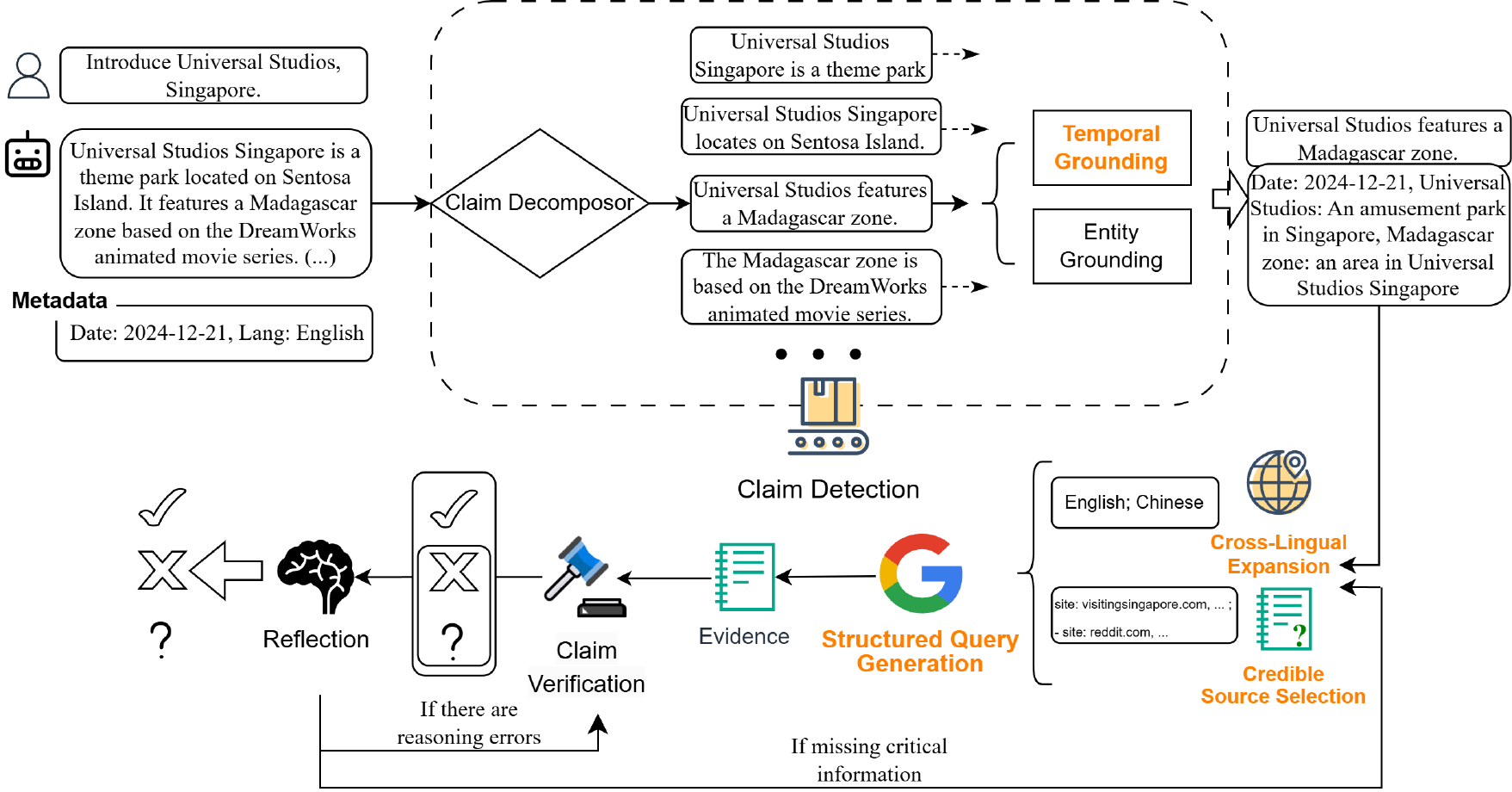}
  \captionsetup{font=small}
  \caption {Overview of the PASS-FC framework. We claim all of the procedures highlighted in \textbf{\textcolor{lightblue}{bold orange}} as our initial contributions to the automatic fact-check (AFC) task.}
  \label{fig:method}
\end{figure*}

\begin{figure}[t] 
\centering 
\begin{subfigure}{\columnwidth} 
\centering 
\includegraphics[width=\columnwidth]{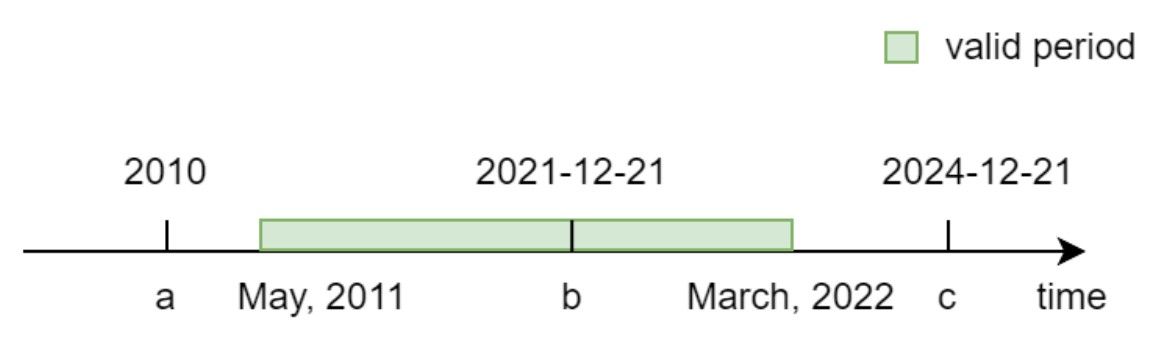} 
\captionsetup{font=scriptsize}
\caption{I. Timeline of the Madagascar zone.  The green bar shows its verified open period obtained from authoritative sources; any claim period outside this bar is unsupported.}  
\label{subfig:claim_time} 
\end{subfigure} 

\begin{subfigure}{\columnwidth} 
\centering 
\footnotesize  
\setlength{\tabcolsep}{4pt}  
\renewcommand{\arraystretch}{1.2}  
\begin{tabular}{p{0.7\columnwidth}|c}  
\toprule
\textbf{Claim} & \textbf{Claim Period} \\ 
\midrule
a. Universal Studios features a Madagascar zone in \textcolor{lightblue}{2010}. & 2010 \\ 
b. Universal Studios features a Madagascar zone \textcolor{lightblue}{three years ago}. & 2021-12-21 \\ 
c. Universal Studios features a Madagascar zone. & 2024-12-21 \\ 
\bottomrule
\end{tabular} 
\captionsetup{font=scriptsize}
\caption{II. Mapping different \textcolor{lightblue}{temporal formulations} (a, b, c) to claim periods given the validation date \textbf{2024-12-21}.}  
\label{subfig:claim_table} 
\end{subfigure} 
\captionsetup{font=small}
\caption{Temporal grounding example for the claim ``Universal Studios features a Madagascar zone.''} 
\label{fig:claim_date} 
\end{figure}

\paragraph{Task formulation.}
Given a user prompt $p$ (in any language) and a free-form answer $r$ produced by either a human or an LLM, the goal is to decide whether the information in $r$ is true, false, or unverifiable.  All judgments are made with respect to $p$ and the accompanying metadata (e.g.\ validation date).

\paragraph{PASS-FC workflow.}
Figure~\ref{fig:method} shows the pipeline.  
(1)~We first split $r$ into a set of \emph{comprehensive claims} $\{c_1,\ldots,c_n\}$.  
(2)~Each claim then enters an iterative loop composed of adaptive search, verification, and reflection (§\ref{sec:querygen}).  
After every round reflection either stops the loop or issues new tool instructions for the next cycle.

\subsection{Claim Detection}
This front-end stage converts a long answer into self-contained atomic facts and augments them with essential context.

\subsubsection{Claim Decomposition}
Following \citet{afacta,molecularf,factscore}, an \emph{atomic fact} is a minimal statement that (i) is objectively checkable, (ii) conveys exactly one proposition, and (iii) remains intelligible without extra context.  
Given $(p,r)$, an LLM decomposes $r$ into $\{c_1,\ldots,c_n\}$ that satisfy the above criteria (prompt in Fig.~\ref{prm:claim_decomp}).

\subsubsection{Contextual grounding.}\label{sec:claimaug}
Even a self-contained claim may hide ambiguities about \emph{when} it holds and \emph{which} real-world entities it mentions.  
Therefore we append two concise fields to every atomic fact:

\paragraph{Entity Grounding} A unique cue phrase or descriptor for each entity referenced. It focuses on uniquely specifying each entity referenced in a claim \citep{qabrief}. Using the given prompt, response, and metadata, a language model reasons to provide concise descriptions for every entity mentioned.


\paragraph{Temporal Grounding}
The goal of temporal grounding is to attach a precise time span—called the \emph{claim period}—to every atomic claim. The claim period is the interval in which the statement must hold for the claim to be true. Our procedure is purposely simple and deterministic, yet robust to the mismatch between the dataset’s validation date and the date on which the LLM is queried.

\begin{enumerate}[label=\textbf{\arabic*.}, leftmargin=*]
  \item Parse temporal cues (see Fig.~\ref{fig:claim_date} for an example):
        \begin{enumerate}[label=\textbf{\alph*)}, leftmargin=1.5em]
          \item \textbf{Absolute description} — if the claim contains an explicit time span (e.g., ``in~2010''), we take that span verbatim as the claim period.
          \item \textbf{Relative description} — if the claim uses a relative phrase (e.g., ``three years ago''), we resolve it with respect to the validation date $t_{\text{val}}$ supplied by the dataset.
          \item \textbf{No description} — if no temporal cue appears, we default to $t_{\text{val}}$ itself.
        \end{enumerate}

  \item Align the LLM’s notion of ``today'':\\
        LLMs are aware of the real calendar date $t_{\text{now}}$, whereas claims were annotated on $t_{\text{val}}$. Querying the model directly at $t_{\text{now}}$ may introduce fact drift. We therefore rewrite $t_{\text{val}}$ as the literal token \texttt{Now} in the prompt, let the model reason under this neutral anchor, and—after generation—replace \texttt{Now} with the real validation date. This single post-processing step decouples inference from wall-clock time and empirically stabilises accuracy.
\end{enumerate}
A claim is deemed supported only if its derived claim period is fully covered by the time span of the retrieved evidence. The procedure grounds every claim to a fixed temporal frame, removes ambiguity for both absolute and relative descriptions, and, importantly, decouples LLM inference from the moving target of the current date—a key step toward robust, time-aware fact checking. The prompt is showed at Figure~\ref{prm:claim_augment}

The final \emph{comprehensive claim} $\langle$atomic fact, entity ground, temporal ground$\rangle$ is the unit that feeds the adaptive search loop.




\subsection{Adaptive Search Scheme}
\label{sec:querygen}
The Adaptive Search Scheme equips PASS-FC with a focused, progressive retrieval mechanism.
Given a comprehensive claim and the full reasoning trace, it produces search queries that maximize both relevance and trustworthiness of the evidence.
At each iteration the reflection module selects the most suitable search tool with the Structured Query Generation (SQG) tool being the default choice for the initial verification attempt.


\subsubsection{Credible-Source Selection}
Credible-Source Selection (CSS) restricts the search space to domains that are likely to host reliable information. Conditioned on the current claim and earlier results, the LLM proposes a concise allow-list and discards sites of dubious quality.
For instance, Figure~\ref{fig:method} shows that \texttt{visitingsingapore.com}—an official portal inferred from previous hits—is retained, while \texttt{reddit.com} is dropped because its content is user-generated.
In contrast to prior work \citep{averitec, folk} that hard-codes a site list, CSS is fully automatic yet remains user-controllable: a user may also inject preferred domains (e.g., \texttt{*.wikipedia.org}).
The generated list is forwarded to SQG through the prompt in Figure~\ref{prm:site}.

\subsubsection{Cross-Lingual Expansion}
\label{sec:xle}
Cross-Lingual Expansion (XLE) widens the evidence net beyond the user-specified source language (English is only the default when the user gives no preference).
Given the current claim and its metadata, XLE selects at most two extra languages from the 46 that Google Search supports\footnote{\url{https://support.google.com/googleplay/android-developer/table/4419860?hl=en&sjid=11904773475773808427-AP}}.
Selection follows three adaptive rules:

\noindent \textbf{Content}.
If the claim discusses a country or culture, XLE adds that locale’s dominant language (e.g., “Singapore”, Malay or Mandarin).

\noindent\textbf{Source}.
If the claim’s provenance hints at another linguistic context—through the URL domain, speaker nationality, publication venue, or any metadata field—XLE adds that language even when the claim itself is written in the source language or its content does not trigger the former rule.

\noindent\textbf{Diversity}.
When the first two rules still leave the evidence sparse, XLE adds one high-resource language that is highly indexed for the topic (e.g., Spanish, French).
If earlier retrieval has already yielded sufficient evidence, this step is skipped.

The source language is always preserved. For each language, XLE calls SQG to generate new language-specific queries (prompt in Figure~\ref{prm:multilingual}).

\subsubsection{Structured Query Generation} 
SQG converts the selected domains, languages, and temporal/entity grounds into executable search strings.
It exploits advanced operators—\texttt{site:}, quotation marks, Boolean connectors, exclusion symbols, wildcards, and parentheses—to sharpen or broaden the search as needed. These operators are specialized syntax employed by various search engines. 
For every claim, SQG emits two complementary queries: one precision-oriented and one recall-oriented.
A complete description of the operator set and the prompt template appears in Figure~\ref{prm:advanced}.




\subsection{Verification, Reflection, and History Management.}
\label{sec:reflect}
For every structured query, we call the Google Search API\footnote{\url{https://serper.dev/}} and keep the top-$k$ results ($k{=}10$, title–snippet–URL).  
The verifier LLM inspects the claim plus this evidence set and returns one of three labels: \textbf{supported} (most credible, temporally aligned evidence agrees with the claim); \textbf{contradicted} (credible,  temporally aligned evidence refutes it); \textbf{inconclusive} (relevant evidence is missing or mutually conflicting).  
After each round, a \emph{reflection} routine decides whether to (i) stop, (ii) launch an additional search via CSS/XLE/SQG, or (iii) re-evaluate the same evidence when the reasoning appears wrong.  
A light-weight \emph{history manager} stores the claim, current evidence, verdict, and feedback.  With long-context models ($>$8k tokens) the full log is preserved; with shorter-context models (e.g., GPT-3.5-Turbo) we keep only the claim, the most recent iteration, and a concise summary of earlier steps.  The loop terminates when reflection signals “stop’’ or when a preset iteration budget is exhausted.

%% file: section/exp.tex
\definecolor{lightgray}{rgb}{0.9,0.9,0.9}

\begin{figure*}[t]
  \includegraphics[width=0.32\linewidth]{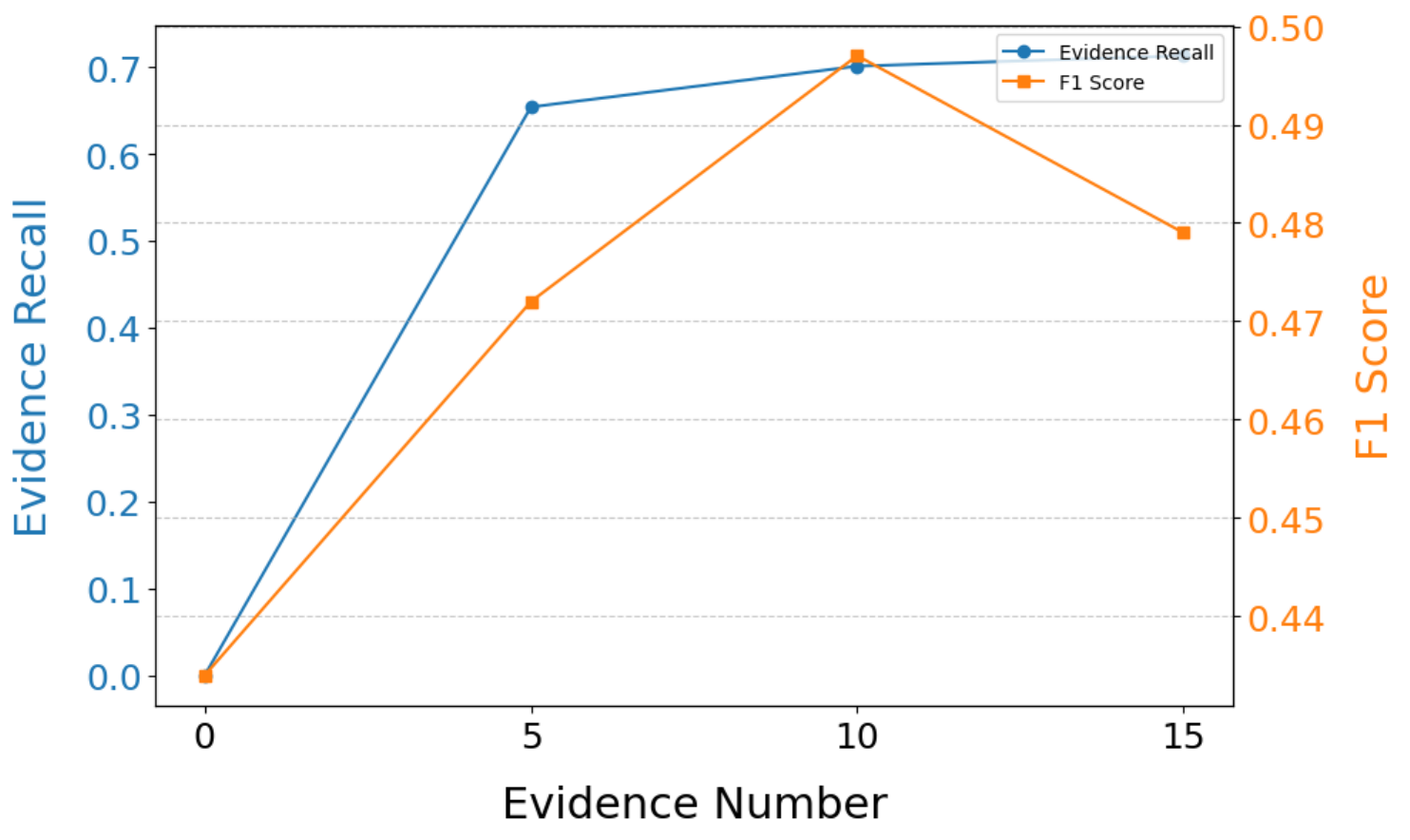} \hfill
  \includegraphics[width=0.32\linewidth]{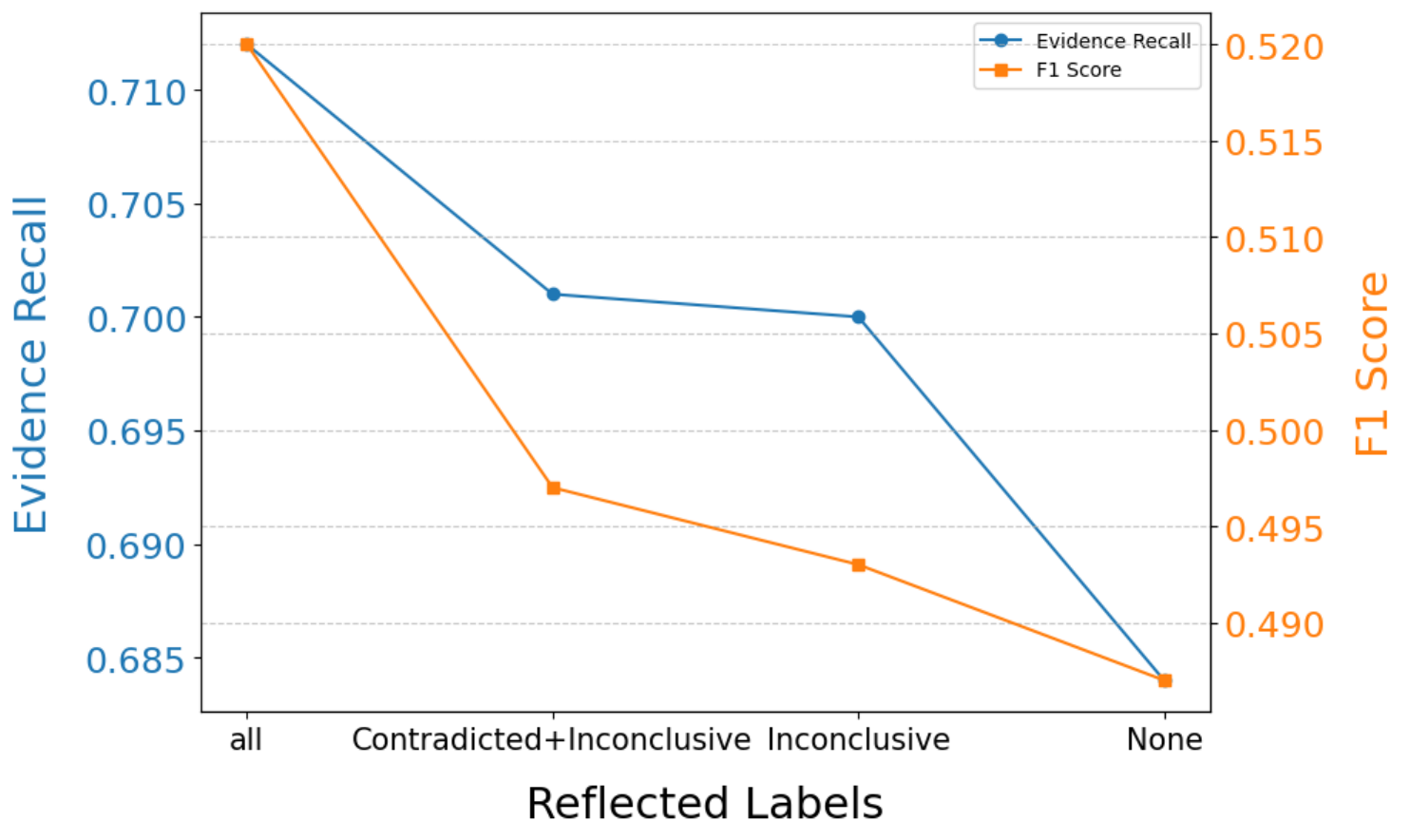} \hfill
  \includegraphics[width=0.32\linewidth]{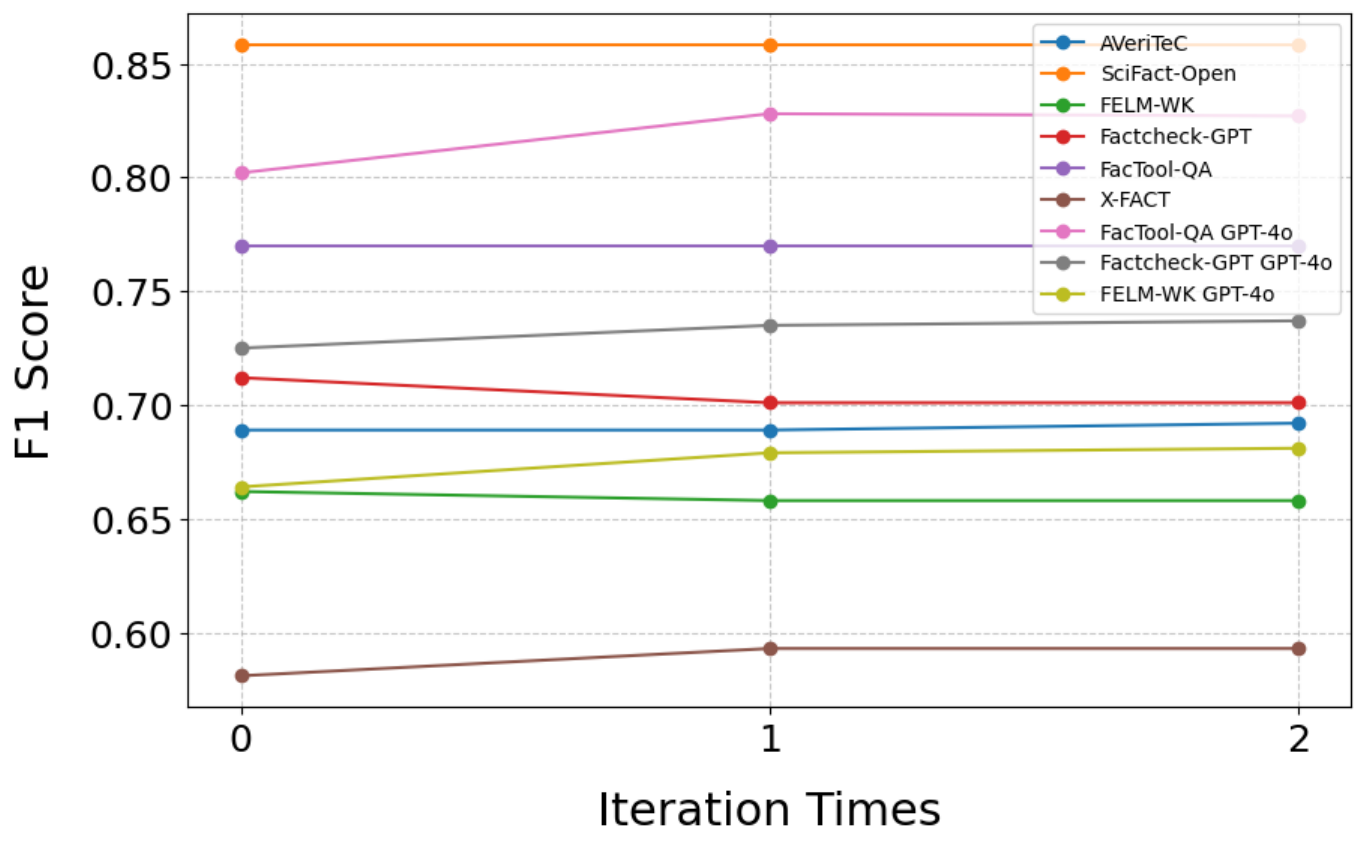}
  \captionsetup{font=small}
  \caption {Hyperparameter analysis. (a) and (b): Impact of evidence number and reflection trigger labels on performance, using 100 randomly sampled AVeriTeC training examples. (c) Performance gains from iterations across all datasets using GPT-4omini, and on Factbench using GPT-4o. Evaluation metrics include evidence recall (where applicable) and macro-F1 score.}
  \label{fig:exp_2}
\end{figure*}

\noindent\textbf{Datasets.} We evaluate PASS-FC on six publicly available collections: FacTool-QA, FELM-WK, Factcheck-GPT, SciFact-Open, AVeriTeC-Dev, and X-FACT \citep{factool,felm,factcheckgpt,wadden2022scifactopenopendomainscientificclaim,averitec,xfactnewbenchmarkdataset}.
Following \citet{openfactcheck}, FacTool-QA, FELM-WK, and Factcheck-GPT are grouped as \emph{FactBench}, a knowledge-oriented QA suite for which human-written atomic claims are already provided; we therefore skip claim decomposition on that subset.
SciFact-Open targets scientific statements, AVeriTeC-Dev covers real-world claims, and X-FACT stresses cross-lingual verification. Detailed descriptions and statistics (Table~\ref{tab:stat}) appear in Appendix~\ref{sec:dataset}.


\noindent\textbf{Preventing data leakage.} AVeriTeC and X-FACT supply a verification date and the URLs of fact-checking sites. At query time we restrict evidence to documents \emph{published before} the verification date, which should block the official fact-checking domains. A faux domains list provided in AVeriTec is also blocked.
The remaining four datasets lack explicit timestamps, so we conservatively use the arXiv publication date of the dataset paper as the cut-off. FactBench and SciFact-Open are not sourced from fact-checking portals, hence pose no additional leakage risk. A discussion on search-engine reproducibility is given in Appendix~\ref{app:search_engine}.


\noindent\textbf{Baselines.} We compare PASS-FC with four recent LLM-based fact-checkers that also rely on Google retrieval: FacTool, Factcheck-GPT, SAFE, and FIRE. Each model runs with the authors’ default hyper-parameters; descriptions (\ref{app:baselinedes}), implementation details (Table~\ref{tab:models_settings}) and our rationale for their selection (\ref{app:rational}) are summarized in Appendix.


\noindent\textbf{Metrics.} Following prior work \citep{folk,selfchecker}, we report accuracy and macro-F1, where macro-F1 averages the class-wise F1 scores of \textit{Supported} and \textit{Contradicted}. Because most datasets do not contain the \textit{Inconclusive} label that PASS-FC can output, this metric setup mildly penalizes our system (See Appendix~\ref{labelstand} for detailed discussion). Aggregate results appear in Table~\ref{tab:main_exp}.


\begin{table*}[t!] 
\scriptsize
    \centering 
    \resizebox{\textwidth}{!}{%
    \begin{tabular}{l|c|cc|cc|cc|cc|cc|cc} 
    \toprule
     & & \multicolumn{2}{c|}{FacTool-QA} & \multicolumn{2}{c|}{FELM-WK} & \multicolumn{2}{c|}{Factcheck-GPT} & \multicolumn{2}{c|}{SciFact-Open} & \multicolumn{2}{c|}{AVeriTeC} &\multicolumn{2}{c}{Average}\\ 
    \multicolumn{1}{c|}{Framework} &  \multicolumn{1}{c|}{Base Model} & F1 & Acc & F1 & Acc & F1 & Acc & F1 & Acc & F1 & Acc & F1 & Acc \\ 
    \midrule 
    Always True & - & 43.2 & 76.0 & 41.5 & 71.0 & 41.0 & 69.6 & 35.6 & 52.9 & 19.6 & 24.4  & 36.2 & 58.8\\
    Always False & - & 19.4 & 24.0 & 22.5 & 29.0 & 19.0 & 23.5 & 32.0 & 47.1 & 37.9 & 61.0 & 26.2 & 36.9\\
    Random Guess & - & 47.3 & 51.5 & 47.3 & 49.9 & 46.6 & 48.1 & 55.0 & 55.0 & 38.2 & 36.8 & 46.9 & 48.3\\
    \midrule 
    FacTool & GPT-3.5-Turbo & \cellcolor{lightgray} 62.5 & \cellcolor{lightgray} 65.0 & \cellcolor{lightgray} 56.0 & \cellcolor{lightgray} 60.6 & \cellcolor{lightgray} 63.5 & \cellcolor{lightgray} 68.6 & 79.6 & 79.6 & 58.2 & 57.0 & 64.0 & 66.2 \\ 
    Factcheck-GPT & GPT-4/GPT-4omini & \cellcolor{lightgray} 75.5 & \cellcolor{lightgray} 81.8 & \cellcolor{lightgray} 66.5 & \cellcolor{lightgray} \textbf{76.3} & \cellcolor{lightgray} 71.0 & \cellcolor{lightgray} 67.4 & 58.8 & 56.0 & 57.6 & 56.2 & 65.9 & 67.5 \\ 
    SAFE & GPT-3.5-Turbo & 59.1 & 62.7 & 46.5 & 46.6 & 57.2 & 55.0 & 67.6 & 68.6 & 61.0 & 63.0 & 58.3 & 59.2\\ 
    FIRE & GPT-4o & \cellcolor{lightgray} \underline{79.0} & \cellcolor{lightgray} \underline{85.1} & \cellcolor{lightgray} \textbf{70.0} & \cellcolor{lightgray} 63.9 & \cellcolor{lightgray} \underline{75.5} & \cellcolor{lightgray} 73.7 &  \textbf{87.4} & \textbf{87.4} & \underline{70.7} & 69.2 & \underline{76.5} & 75.9 \\
    FIRE & GPT-4omini & \cellcolor{lightgray} 73.0 & \cellcolor{lightgray} 80.9 & \cellcolor{lightgray} 62.0 & \cellcolor{lightgray} 53.0 & \cellcolor{lightgray} \textbf{77.0} & \cellcolor{lightgray} \textbf{76.1} & 78.9 & 79.1 & 69.6 & 69.2 & 72.1 & 71.7 \\
    \midrule
    PASS-FC & GPT-3.5-Turbo & 67.2 & 78.5 & 58.6 & 70.4 & 69.8 & \underline{75.7} & 72.8 & 73.3 & 66.6 & 65.8 & 67.0 & 72.7\\ 
    PASS-FC & GPT-4o &  \textbf{82.8} & \textbf{86.3} & \underline{68.2} & \underline{71.2} & 74.0 & 74.6 & 85.7 & 82.2 & \textbf{77.6} & \textbf{72.0} & \textbf{77.7} & \textbf{77.3} \\
    PASS-FC & GPT-4omini & 77.0 & 81.1 & 66.5 & 70.0 & 70.9 & 72.6 & \underline{85.8} & \underline{85.9} & 69.9 & \underline{70.2} & 74.0 & \underline{76.0}\\ 
    \midrule
- Temporal Grounding & GPT-4omini & 73.8 & 77.3 & 65.3 & 68.4 & 70.9 & 71.8 & 84.3 & 84.2 & 68.3 & 68.8 & 72.5 & 74.1\\
- Structured Query Generation & GPT-4omini & 75.2 & 78.5 & 65.3 & 67.8 & 71.2 & 72.3 & 85.3 & 85.3 & 68.7 & 69.6 & 73.1 & 74.7\\
- Credible-Source Selection & GPT-4omini & 74.9 & 78.3 & 66.2 & 69.2 & 70.4 & 72.3 & 82.1 & 82.2 & 69.3 & 69.8 & 72.6 & 74.4\\
- Cross-Lingual Expansion & GPT-4omini & 77.0 & 81.1 & 66.2 & 69.6 & 70.9 & 72.6 & 85.8 & 85.9 & 69.2 & 69.4 & 73.8 & 75.7 \\
+ force XLE & GPT-4omini & 78.7 & 83.3 & 67.1 & 71.8 & 71.9 & 73.3 & 86.9 & 86.9 & 72.8 & 71.2 & 75.5 & 77.3 \\
    \bottomrule
    \end{tabular}} 
    \captionsetup{font=small}
    \caption{Macro-F1 and accuracy for the fact-checking task. The highest results are in \textbf{bold}, and the second-best results are \underline{underlined}. Results within the shaded area are cited from \citet{fire} or \citet{openfactcheck}, where Factcheck-GPT uses GPT-4-Turbo as the base model. We ran the remaining Factcheck-GPT results with GPT-4omini to limit costs. Although FIRE used Factcheck-GPT as their development set, we still retain its result here for completeness.} 
    \label{tab:main_exp} 
\end{table*}
\noindent\textbf{PASS-FC configuration.}
To ensure fair cross-dataset comparison, no parameter is tuned on any test split.
Instead, we probe the main hyper-parameters on 100 randomly sampled AVeriTeC-train instances with \texttt{GPT-4o-mini}.
Figure \ref{fig:exp_2}a varies the number of snippets kept per query: recall climbs steadily, but macro-F1 peaks at ten snippets and drops thereafter as noise accumulates.
Figure \ref{fig:exp_2}b alters the set of verdicts that trigger reflection: performance improves whenever more labels are allowed to relaunch the search, confirming the utility of the reflection loop.

Guided by these trends we freeze one global setting for all subsequent experiments:
English is taken as the source language (X-FACT uses its own language metadata); all adaptive search tools remain enabled; we keep the top-10 hits for each query; the loop is capped at two iterations; reflection fires only on \textit{Contradicted} or \textit{Inconclusive}.
Results are reported for \texttt{GPT-3.5-Turbo}, \texttt{GPT-4o}, and \texttt{GPT-4o-mini}, each run with temperature = 0.01.



\subsection{Main results}\label{exp:exp1}
Table~\ref{tab:main_exp} compares PASS-FC with state-of-the-art few-shot fact-checkers. Token costs on AVeriTeC-Dev and SciFact-Open are presented in Table~\ref{tab:costs}.
Two trends are evident.  

Firstly, PASS-FC is highly competitive even when it runs on weaker back-ends.  
With \texttt{GPT-3.5-Turbo} it already surpasses all baselines that use the same model.  
With \texttt{GPT-4o-mini} it matches or exceeds \textsc{Factcheck-GPT} and delivers higher average accuracy than \textsc{FIRE} (which in turn relies on a larger \texttt{GPT-4o}) despite FIRE being tuned on Factcheck-GPT testset.  
Using \texttt{GPT-4o}, PASS-FC attains the best mean F1 and accuracy overall.

Secondly, performance is stable across domains.  
The framework not only excels on the knowledge-oriented FactBench suite but also on SciFact-Open (scientific claims) and AVeriTeC (time-stamped real-world claims), where it outperforms every baseline by a wide margin—reflecting the benefit of explicit temporal grounding.  
Similar behaviour is observed on the multilingual X-FACT benchmark (Table~\ref{tab:xfact_res} in the appendix): PASS-FC outperforms every baseline. Furthermore, for all frameworks, accuracy is highest for languages typologically close to English (e.g.\ Norwegian) and lowest for distant languages such as Serbian, mirroring the pattern reported by \citet{huang-etal-2022-concrete}.  
We also compare PASS-FC with models introduced in the AVeriTeC Challenge \citep{schlichtkrull-etal-2024-automated} in table~\ref{tab:averitec_comp} from \ref{app:averitec_text}. PASS-FC is also excellent when tested under the same setting.

Figure~\ref{fig:exp_2}c studies the effect of reflection iterations in the main experiments.  
\texttt{GPT-4o} continues to improve when allowed a second pass, whereas \texttt{GPT-4o-mini} gains little, confirming that stronger models can exploit the reflection loop more effectively.

\begin{figure*}[t]
    \centering
  \includegraphics[scale=0.3]{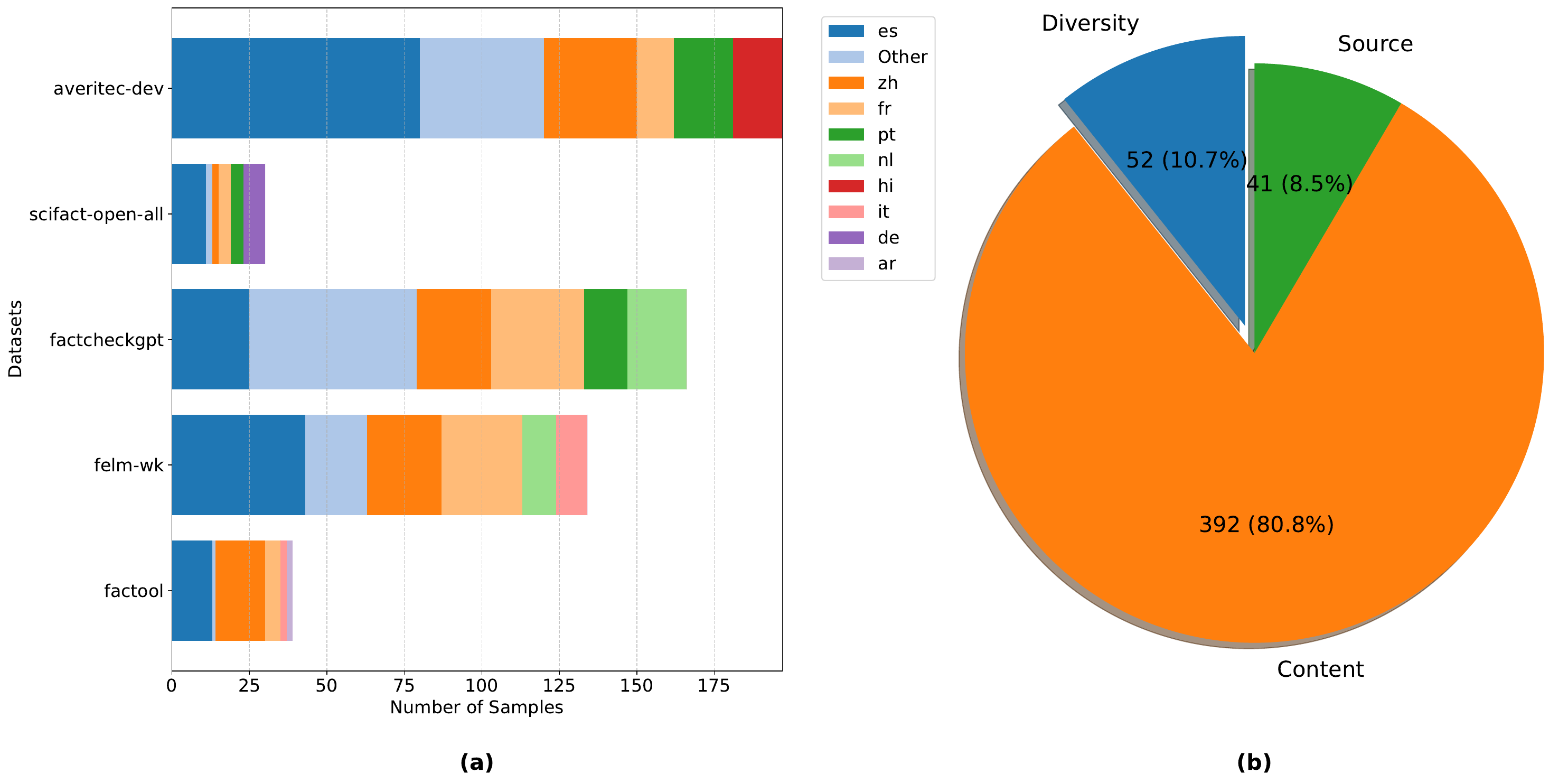}
  \captionsetup{font=small}
  \caption {Analysis of XLE. (a): Distribution of languages chosen by XLE in the \textit{force-XLE} setting from the ablation study. (b) The distribution of reasons that choose every language.}
  \label{fig:mul_dist}
\end{figure*}

\subsection{Ablation study}\label{exp:exp3}
The lower block of Table~\ref{tab:main_exp} removes each of the four novel components in turn while keeping \texttt{GPT-4o-mini} and all other hyper-parameters fixed.

Temporal grounding has the largest single impact (-2.0 accuracy on average); even the coarse cut-off date “2023-07-26“ used for the entire FactTool-QA set prevents many stale assertions from being judged correct.  
Replacing structured queries with plain natural-language questions also degrades F1, showing that advanced operators are essential for precise retrieval.  
Credible-source selection mainly helps on SciFact-Open, where model-selected peer-reviewed and governmental domains are crucial for reliable scientific evidence.

Cross-lingual expansion (XLE) fires under our default, conservative setting in only 5$\%$ of cases, so turning it off has little effect.  
Forcing XLE to run whenever a new iteration is triggered boosts average F1 by 1.5 points at the cost of roughly 10 $\%$ extra runtime, suggesting that wider language coverage can pay off when latency permits.  
On X-FACT (Table~\ref{tab:xfact_res} in the appendix) the benefit is much larger because English, now treated as a foreign but high-resource language, supplies abundant evidence, especially for Russian, Turkish, and Portuguese claims.  
Conversely, asking the model to \emph{reason} in the source language rather than English hurts performance for low-resource languages—reasoning quality evidently depends on the language distribution in the model’s training data.

\begin{table}[h]
\centering
\scriptsize
\begin{tabular}{lcc}
\toprule
Testsets      & Query Similarity & Evidence Similarity \\
\midrule
FacTool-QA    & 91.2             & 85.6                \\
FELM-WK       & 90.6             & 84.2                \\
Factcheck-GPT & 91.5             & 83.3                \\
SciFact-Open  & 92.8             & 86.9                \\
AVeriTeC      & 90.5             & 83.4                \\
\bottomrule
\end{tabular}
\caption{Mean cosine similarity (\%) between (i) XLE queries and all other queries, and (ii) XLE evidence and all other evidence, measured with the multilingual \textsc{e5} encoder.}
\label{tab:mul_dist_sim}
\end{table}

\subsection{Analysis of XLE}
\label{exp:xle_analysis}
We examine the \textit{force-XLE} setting from the ablation study to understand how multilingual retrieval aids the overall pipeline.

\paragraph{Language distribution.}
Figure~\ref{fig:mul_dist} plots the languages chosen by XLE when it is always activated (top-4 languages are displayed). Spanish, Chinese, and French dominate, matching their global web presence.

\paragraph{Why each language?}
For every XLE we asked GPT-4o to label the reason as \emph{Content}, \emph{Source}, or \emph{Diversity} (§\ref{sec:xle}). A manual audit of 100 samples confirmed $\!>\!90\%$ precision.  Content accounts for the largest share, while Source and Diversity contribute roughly equally.  AVeriTeC is the only benchmark that provides URL, location, and speaker metadata. Source-based triggers are common—explaining the large gain from XLE on this set.

\paragraph{Complementary evidence.}
Does XLE merely duplicate English evidence or does it add new information?  
Using the official AVeriTeC gold evidence we found \emph{no} multilingual snippet with cosine similarity $>0.85$ to any English gold snippet. Qualitative inspection shows that XLE often supplies parallel reports in local media (eg. the case study in Figure~\ref{case:xle}), furnishing an alternative validation path—a pattern also observed in the AVeriTeC human-study where pairwise trace overlap is below 0.25 even within the same language.

Table~\ref{tab:mul_dist_sim} extends the analysis to all five test sets. While XLE queries are highly similar to the monolingual ones ($\ge 90\%$), the retrieved documents are not (all $<86\%$), indicating that the search engine’s language alignment plus the heterogeneous web bring genuinely fresh evidence into the loop.

\subsection{Case Study}
\label{exp:case_analysis}
To highlight the internal dynamics of PASS-FC, we walk through two concise examples in Appendix~\ref{app:case} (Figures~\ref{case:tg} and \ref{case:xle}) that isolate the roles of Temporal Grounding and XLE.

In the first case, PASS-FC first attaches the year ‘‘2023,’’ yielding the grounded claim \emph{In~2023, the United States has 94 operating nuclear reactors.''} This single token changes the outcome. In 2024 the U.S.\ \emph{does} list 94 active reactors,\footnote{\url{https://www.eia.gov/tools/faqs/faq.php?id=207&t=21}} so a system that searches immediately is likely to retrieve supportive snippets and label the claim \textsc{Supportive}. Grounding to 2023 steers the search toward historical statistics, from which PASS-FC retrieves evidence that only 93 reactors were operational that year,\footnote{\url{https://www.eia.gov/energyexplained/nuclear/us-nuclear-industry.php}} and outputs \textsc{Contradicted}.

The second example starts with an English-only search that returns conflicting evidence, causing the first verification attempt to stall. In the next round, XLE produces Spanish and Portuguese query variants. The Spanish query surfaces a CNN~en Español article that squarely addresses the claim. This new evidence lets PASS-FC make the right verdict, illustrating how multilingual retrieval supplies otherwise unreachable support.



\section{Conclusion}
We introduced PASS-FC, a fact-checking framework that augments large language models with temporal grounding and an adaptive, progressively refined search scheme. Across six benchmarks—spanning general knowledge, scientific claims, real-world events, and ten languages—PASS-FC outperforms strong baselines even when running on smaller backbone models. As on-line information continues to proliferate and shift over time, methods such as PASS-FC that explicitly reason about time, provenance, and language will be essential for maintaining information integrity and supporting reliable, automated decision making.

%% file: section/Appendix.tex
\UseRawInputEncoding
\appendix

\section{Appendix}
\label{sec:appendix}

\subsection{Dataset Descriptions and Statistics}
\label{sec:dataset}
\paragraph{Factool-QA} The Factool-QA dataset \citep{factool} is a collection designed for fact-checking and question-answering tasks. It contains 50 real-world questions along with responses generated by ChatGPT. Each response is accompanied by annotated claims extracted from the AI-generated content. The questions in this dataset are sourced from various platforms, including Quora and TruthfulQA, providing a diverse range of topics. This dataset serves as a valuable resource for evaluating the accuracy of AI-generated responses and testing fact-checking systems.

\paragraph{FELM-WK} The FELM-WK (World Knowledge) dataset \citep{felm} is a subset of the larger FELM (Factuality Evaluation for Large language Models) collection, specifically focused on world knowledge. It contains 184 examples, which are further divided into 532 sub-claims. This dataset covers a wide range of topics including history, society, common sense, and current events. The questions in FELM-WK are sourced from various datasets and platforms such as TruthfulQA \citep{lin2022truthfulqameasuringmodelsmimic}, Quora, hc3 \citep{guo2023closechatgpthumanexperts}, and MMLU \citep{lin2022truthfulqameasuringmodelsmimic}, as well as some questions generated by ChatGPT and curated by the authors. FELM-WK is designed to evaluate the factual accuracy of language models in generating responses related to general world knowledge, making it a valuable resource for testing fact-checking systems and assessing the factual reliability of AI-generated content.

\paragraph{FactCheckGPT} The FactCheckGPT dataset \citep{factcheckgpt} is a comprehensive collection designed for evaluating fact-checking systems and language models. It contains 184 examples, which are further divided into 532 sub-claims. The dataset focuses on fact-intensive content where language models are prone to hallucinate or produce factual errors. The examples are sourced from various origins, including ChatGPT-generated responses posted on Twitter, in-house brainstorming sessions, and selections from the dolly-15k dataset \citep{DatabricksBlog2023DollyV2}. The claims in FactCheckGPT are annotated for importance and checkworthiness, making it a valuable resource for testing fact-checking methodologies and assessing the factual accuracy of AI-generated content across a range of topics and complexities.

\paragraph{Scifact-Open} The SciFact-Open dataset \citep{wadden2022scifactopenopendomainscientificclaim} is distinct from previously mentioned datasets in that its claims are not generated by language models but are derived from scientific literature. In our study, we utilize a subset of 191 claims from SciFact-Open that have factuality labels. Unlike the original dataset design, which includes a corpus of abstracts for verification, our approach diverges by using Google Search for evidence retrieval instead of the provided evidence set. This modification allows us to test our fact-checking framework in a more open-ended, web-based setting. The SciFact-Open claims, being grounded in scientific literature, provide a rigorous benchmark for evaluating fact-checking systems on specialized, technical content, particularly in the domains of medicine and biology.

\paragraph{AVeriTeC} The AVeriTeC (Automated VERIfication of TExtual Claims) dataset \citep{averitec} is a comprehensive resource for fact-checking research. In our study, we utilize the development set of AVeriTeC, as the test set labels are not publicly available. The dataset originally contains real-world claims annotated with question-answer pairs, veracity labels, and textual justifications. For our purposes, we modified the labeling scheme by merging the "not enough evidence" and "conflicting evidence/cherry-picking" categories into a single "inconclusive" label, aligning with our framework's classification approach. This adaptation of AVeriTeC provides a challenging benchmark for evaluating our fact-checking system on real-world claims, while maintaining a three-class labeling system (supported, refuted, inconclusive) consistent with our research objectives.

\paragraph{X-FACT} 
\label{app:xfact_constuction}
The X-FACT dataset \cite{xfactnewbenchmarkdataset}, originally a multilingual fact-checking resource, has been adapted for our study. We selected claims labeled as either True or False, excluding other categories. From this subset, we chose 10 languages where both True and False labels had more than 50 instances each. These languages include English, Spanish, Italian, Indonesian, Polish, Portuguese, Romanian, Serbian, Turkish, and Russian. For each language, we randomly sampled 50 claims per label. The first eight languages are used from the training set, while Turkish and Russian serve as zero-shot test languages. This modification allows us to evaluate our fact-checking system's performance across diverse languages.

\subsubsection{Statistics}
Detailed distribution are presented in Table~\ref{tab:stat}.

\begin{table*}[h]
\centering
\small
\begin{tabular}{l|ccccc}
\toprule
\textbf{Datasets}       & \textbf{Supported} & \textbf{Contradicted} & \textbf{Inconclusive} & \textbf{Conflicting/Cherrypicking} & \textbf{All} \\
\midrule
FacTool-QA    & 177 (76.0\%)  & 56 (24.0\%)    & N/A          & N/A                       & 233  \\
FELM-WK       & 360 (71.0\%) & 147 (29.0\%)  & N/A          & N/A                       & 507  \\
Factcheck-GPT & 472 (69.6\%) & 159 (23.5\%)  & 47 (6.9\%)    & N/A                       & 631  \\
Scifact-open  & 101 (52.9\%) & 90 (47.1\%)   & N/A          & N/A                       & 191  \\
AVeriTeC-Dev  & 122 (24.4\%) & 305 (61.0\%)  & 35 (7.0\%)    & 38 (7.6\%)                 & 500  \\
X-FACT        & 500 (50.0\%) & 500 (50.0\%)  & N/A          & N/A                       & 1000 \\
\bottomrule
\end{tabular}
\caption{Datasets distributions. X-FACT is we random sampled balanced testset, with 10 languages, one containing 100 examples. For detailed description, refer to \ref{app:xfact_constuction}}.
\label{tab:stat}
\end{table*}

\subsection{Impact of Search Engines}
\label{app:search_engine}
Google search results may change over time. This issue has been studied in fact-checking research: \citet{wildevd} shows that while only 30\% of search output URLs overlap when queried two months apart, the veracity judgment classification remains largely stable. We've verified this in our experiments. Rerun our main experiment after a two-month interval on FactCheck-GPT only slightly change the result (within 0.5\% on accuracy), demonstrating minimal changes from the original results. Using search engines is unavoidable for real-world fact-checking, as they represent one of the largest, continuously updated knowledge base. We consider this a limitation of the approach rather than a weakness, given the nature of real-world fact-checking tasks.

\subsection{Baselines}

\subsubsection{Descriptions}
\label{app:baselinedes}
\paragraph{FacTool} Factool \citep{factool} employs a tool-augmented approach for fact-checking. It operates in five stages: claim extraction using ChatGPT, query generation for each claim, evidence collection via Google Search API, and agreement verification using ChatGPT or GPT-4. This framework integrates large language models with external tools to assess the factuality of claims, providing a comprehensive baseline for comparison with our PASS-FC model.

\paragraph{FactCheckGPT} FactCheckGPT \citep{factcheckgpt} is a comprehensive baseline model that approaches fact-checking through a fine-grained, eight-step process. These steps include decomposition, decontextualization, checkworthiness identification, evidence retrieval and collection, stance detection, correction determination, claim correction, and final response revision. This structured approach allows for a detailed evaluation of each component in the fact-checking pipeline. While designed as a comprehensive framework, it offers flexibility in implementation, allowing for the combination of certain steps in practical applications. FactCheckGPT serves as a detailed comparison point for our PASS-FC model, providing insights into the performance of individual fact-checking subtasks.

\paragraph{SAFE} SAFE (Search-Augmented Factuality Evaluator) \citep{safe} is a baseline model that employs a language model to evaluate the factuality of long-form responses. It operates in three main steps: (1) splitting the response into individual self-contained facts, (2) determining the relevance of each fact to the original prompt, and (3) verifying the factuality of relevant facts through iterative Google Search queries. SAFE's key innovation lies in its use of a language model to generate multi-step search queries and reason about the search results. The model outputs metrics including the number of supported, irrelevant, and not-supported facts. This approach provides a comprehensive factuality assessment, serving as a strong baseline for comparison with our PASS-FC model.

\paragraph{FIRE} FIRE is a fact-checking framework that iteratively integrates evidence retrieval and claim verification. It decides whether to provide a final answer or generate a new search query based on confidence in the current judgment. Notably, FIRE can make a verdict entirely based on the model’s confidence, even without external evidence.

\begin{table}[h]
\scriptsize
\centering
\begin{tabular}{ll}
\toprule
\textbf{Base Model}            & \textbf{Frameworks} \\
\midrule
gpt-4-turbo-2024-0409 & PASS-FC, FactcheckGPT, FIRE \\
gpt-35-turbo-0613     & PASS-FC, FacTool, SAFE \\
gpt-4omini-0718       & PASS-FC, FactcheckGPT, FIRE \\
\bottomrule
\end{tabular}
\caption{Frameworks and tested base models.}
\label{tab:models_settings}
\end{table}

\subsubsection{Settings}
All compared frameworks use the same version of their default base models, as detailed in the following table~\ref{tab:models_settings}. Plus, all baseline framework settings (for instance, temperature and evidence number) used their default settings in their official codebase.

\subsubsection{Rationals for Selecting Baseline Frameworks}
\label{app:rational}

We adopt three criteria when choosing baselines:
(i) the system must target \emph{general-purpose, textual} AFC;
(ii) it must retrieve evidence through a public search engine rather than a closed knowledge base;
(iii) it better includes a reflection module or non-trivial search heuristics, so that a fair comparison with PASS-FC is possible.
All baselines we finally report satisfy these requirements.

Among the candidates in Table~\ref{tab:model_comparison}, Self-Checker \citep{selfchecker} nominally meets the rules, but its public repository offers three distinct prompt sets—one per benchmark—making it unclear which configuration reflects real-world usage; we therefore exclude it. ProgramFC \citep{programfc} and KnowHalu \citep{knowhalu} rely on Wikipedia alone and are thus disqualified by criterion~(ii). PACAR \citep{zhao-etal-2024-pacar-automated} does not release code, preventing reproducibility.

The remaining candidates are HISS \citep{hiss}, VERISCORE \citep{veriscore}, and DEFAME \citep{defame}.
HISS is tailored to news-domain claims and its prompts hard-code that domain, so we omit it.
VERISCORE extends SAFE \citep{safe} with an upfront “trustworthy-claim’’ filter; because all our test sets contain only claims that \emph{require} verification, this extra step is irrelevant, and we adopt the simpler SAFE.
DEFAME, proposed for the AVeriTeC challenge \citep{schlichtkrull-etal-2024-automated}, is the only truly general-purpose system in that competition in our knowledge, but it is designed for mixed text-image claims and is extremely resource-intensive (around 5 h and 50\$ for 200 purely textual samples). We therefore omit it from the main comparison, although we do report a head-to-head result on AVeriTeC-DEV in the next section.

\subsection{Label Standardization}
\label{labelstand}
Most test datasets have binary labels: supported and contradicted. FactCheckGPT and AVeriTeC include an additional 'inconclusive' label, but it only accounts for about 10\% of the claims. Baseline models primarily output 'supported' and 'contradicted' labels. While SAFE can output an 'irrelevant' category, it's rarely used due to the factual nature of the test data.
Thus a more favorable approach for baseline models was adopted. Macro-F1 is calculated as the average of F1 scores for 'supported' and 'contradicted' categories only. Accuracy is computed across all categories, providing a more balanced evaluation.

\definecolor{lightgray}{rgb}{0.9,0.9,0.9}
\begin{table*}[h]
\centering
\begin{tabular}{l|l|l|cc}
\toprule
\textbf{Frameworks} & \textbf{Base Model} & \textbf{Source} & \textbf{F1} & \textbf{Accuracy} \\
\midrule
DEFAME & GPT-4o-0806 & AVeriTeC KB & N/A & \cellcolor{lightgray} 74.0 \\
DEFAME & GPT-4omini-2024-07-18 & AVeriTeC KB & N/A & \cellcolor{lightgray} 59.2 \\
Papelo (first 200) & GPT-4o-0513 & google serper & \cellcolor{lightgray} 77.9 & \cellcolor{lightgray} 75.0 \\
PASS-FC & GPT-4o-0806 & google serper & 77.6 & 72.0 \\
PASS-FC & GPT-4omini-2024-07-18 & google serper & 69.9 & 70.2 \\
PASS-FC & GPT-35-turbo-0613 & google serper & 66.6 & 65.8 \\
DEFAME (first 220) & GPT-4o-0806 & google serper & 71.0 & 61.1 \\
PASS-FC (first 220) & GPT-4o-0806 & google serper & 72.6 & 67.3 \\
\bottomrule
\end{tabular}
\caption{Performance metrics across different framework and model combinations on AVeriTeC-Dev. Results within the shaded area are cited from their original paper \citep{defame, papelo}. Note that Papelo only report results on the first 200 sample of AVeriTeC-Dev and we put the best result amoung their 9 settings here (They treat AVeriTeC-Dev as the development set.)}
\label{tab:averitec_comp}
\end{table*}

\subsection{AVeriTeC Challenge Frameworks Comparison}
\label{app:averitec_text}
Most systems in the AVeriTeC Challenge used AVeriTeC's official knowledge base, which includes all golden evidence plus 1000 irrelevant pieces per claim, simplifying the task. The Papelo work \citep{papelo} is one of only three that used Google search directly (also the best performing one amoung the three), but it is explicitly stated by its authors\footnote{\url{https://github.com/cdmalon/fever2024}} to be specifically designed for AVeriTeC, not for general fact-checking. This situation remains for other frameworks except DEFAME. Therefore, we did not include them in our main experiment comparisons as our focus is on general-purpose fact-checking systems.

We report comparisons with DEFAME and Papelo in Table~\ref{tab:averitec_comp}. DEFAME's performance varies significantly across base models, suggesting heavy reliance on model capabilities. We tested DEFAME with Google search on only the first 200 examples (cost around 5 hours and 50\$). Even with this limited sample, our system outperformed DEFAME using the same search source. Papelo works better on their reported result. But they treat AVeriTeC-Dev as the development set. We put the best result amoung their 9 settings here.

\begin{table*}[t]
\scriptsize
\centering
\setlength{\tabcolsep}{3.25pt}
\begin{tabularx}{\textwidth}{@{}l*{10}{rr}rr@{}}
\toprule
\multirow{2}{*}{Framework} & \multicolumn{2}{c}{no} & \multicolumn{2}{c}{ru} & \multicolumn{2}{c}{es} & \multicolumn{2}{c}{it} & \multicolumn{2}{c}{id} & \multicolumn{2}{c}{pl} & \multicolumn{2}{c}{pt} & \multicolumn{2}{c}{ro} & \multicolumn{2}{c}{sr} & \multicolumn{2}{c}{tr} & \multicolumn{2}{c@{}}{Overall} \\
\cmidrule(lr){2-3} \cmidrule(lr){4-5} \cmidrule(lr){6-7} \cmidrule(lr){8-9} \cmidrule(lr){10-11} \cmidrule(lr){12-13} \cmidrule(lr){14-15} \cmidrule(lr){16-17} \cmidrule(lr){18-19} \cmidrule(lr){20-21} \cmidrule(lr){22-23}
 & F1 & Acc & F1 & Acc & F1 & Acc & F1 & Acc & F1 & Acc & F1 & Acc & F1 & Acc & F1 & Acc & F1 & Acc & F1 & Acc & F1 & Acc \\
\midrule
FacTool & 62.0 & 62.0 & 57.7 & 57.0 & 56.5 & 58.0 & 52.8 & 52.0 & 68.4 & 66.0 & 51.2 & 54.0 & 54.8 & 55.0 & 53.3 & 55.0 & 51.0 & 55.0 & 55.3 & 58.0 & 56.9 & 57.2 \\
FIRE-4o & 65.2 & 67.0 & 62.2 & 64.0 & 56.6 & 60.0 & 47.0 & 48.0 & 73.8 & 74.0 & 60.4 & 64.0 & 56.9 & 58.0 & 56.3 & 58.0 & 53.6 & 57.0 & 58.0 & 60.0 & 59.1 & 61.0 \\
FIRE-4omini & 64.8 & 67.0 & 60.1 & 63.0 & 49.5 & 54.0 & 51.3 & 52.0 & 73.3 & 73.0 & 60.4 & 64.0 & 51.9 & 52.0 & 62.8 & 63.0 & 50.8 & 53.0 & 59.1 & 63.0 & 58.9 & 60.4 \\
passfc-35 & 81.9 & 82.0 & 66.6 & 67.0 & 56.7 & 59.0 & 53.1 & 54.0 & 73.0 & 73.0 & 53.7 & 55.0 & 58.0 & 58.0 & 54.9 & 55.0 & 53.6 & 57.0 & 60.8 & 61.0 & 61.6 & 62.1 \\
passfc-4o & 80.7 & 81.0 & 75.2 & 76.0 & 63.9 & 66.0 & 54.6 & 55.0 & 76.9 & 77.0 & 66.9 & 68.0 & 60.8 & 61.0 & 40.7 & 45.0 & 52.4 & 57.0 & 59.1 & 63.0 & 63.7 & 64.9 \\
passfc-4omini & 80.3 & 81.0 & 66.9 & 69.0 & 52.5 & 58.0 & 46.3 & 48.0 & 70.5 & 71.0 & 63.0 & 67.0 & 65.8 & 66.0 & 46.4 & 50.0 & 39.9 & 49.0 & 54.0 & 60.0 & 59.3 & 61.9 \\
\midrule
- XLE & 77.1 & 78.0 & 58.2 & 62.0 & 51.8 & 58.0 & 49.5 & 51.0 & 70.3 & 71.0 & 60.7 & 65.0 & 63.3 & 64.0 & 46.5 & 51.0 & 40.9 & 49.0 & 47.6 & 56.0 & 57.4 & 60.5 \\
- lang adap & 78.5 & 79.0 & 67.6 & 69.0 & 57.4 & 61.0 & 52.4 & 53.0 & 69.8 & 70.0 & 64.0 & 67.0 & 56.0 & 57.0 & 47.9 & 52.0 & 46.2 & 54.0 & 59.9 & 64.0 & 60.6 & 62.6 \\
\midrule
Average & 73.8 & 74.6 & 64.3 & 65.9 & 55.6 & 59.3 & 50.9 & 51.6 & 72.0 & 71.9 & 60.0 & 63.0 & 58.4 & 58.9 & 51.1 & 53.6 & 48.6 & 53.9 & 56.7 & 60.6 & - & - \\ 
\bottomrule
\end{tabularx}
\caption{Framework performance across different languages on X-FACT. "XLE" and "lang adap" are short for "Cross-Lingual Expansion" and "language adaptation". The last row shows the average result of every column.}
\label{tab:xfact_res}
\end{table*}

\begin{table*}[h]
\centering
\begin{tabular}{l|l|r}
\toprule
\textbf{Framework} & \textbf{Base Model} & \textbf{Money (\$)} \\
\midrule
FacTool & gpt-35-turbo-0613 & 4.31 \\
SAFE & gpt-35-turbo-0613 & 2.89 \\
FactCheck-GPT & gpt-4omini-0718 & 2.31 \\
FIRE & gpt-4omini-0718 & 0.27 \\
FIRE & gpt-4o-0806 & 4.55 \\
PASS-FC & gpt-4o-0806 & 4.74 \\
PASS-FC & gpt-4omini-0718 & 0.19 \\
PASS-FC & gpt-35-turbo-0613 & 0.58 \\
\bottomrule
\end{tabular}
\caption{Total API costs in USD for LLM token consumption across different framework and model combinations. The monetary values represent the total cost of API calls measured during experimental runs on AVeriTeC-Dev and ScirFact-Open.}
\label{tab:costs}
\end{table*}

\subsection{Case Study}
\label{app:case}

We provide two compact examples—one for Temporal Grounding (Figure~\ref{case:tg}) and one for XLE (Figure~\ref{case:xle})—to illustrate how PASS-FC works in practice. In the second example, XLE generates Spanish and Portuguese queries that surface new evidence, altering the final verdict. To save space, we keep only the essential steps (e.g., the evidence passages actually used in validation). Full, unabridged traces are available in the supplementary material along with all experimental results.

\begin{figure*}[t]
\begin{lstlisting}[basicstyle=\ttfamily\scriptsize, frame=single, breaklines=true]
"step_num": 1,
"action": "contextual_grounding",
"original_claim": "The United States has 94 operating reactors"
"augmented_claim": "The United States has 94 operating reactors.\nBackground Info of the Claim (The 'time' key represents the temporal context or validity period of the claim. And description of the entities in the claim is for better understanding of the fact in the claim.)\n\ttime:2023-07-26"

"step_num": 2,
"action": "generate_query",
"result": {
"advanced query": [
    "United States nuclear reactors count 2023", "How many operating nuclear reactors are in the United States?"]}

"step_num": 3
"action": "EvidenceSearch",
"result": {
"United States nuclear reactors count 2023": [
"title: U.S.: number of nuclear power reactors 2023 | Statista\nsnippet: Number of nuclear power reactors in the United States for selected years from 1957 and 2023 Characteristic Number of operable units 2023 93 2022 92 2021 93 2020 94",
"title: U.S. nuclear industry - U.S. Energy Information Administration (EIA)\nAs of August 1, 2023, the United States had 93 operating commercial nuclear reactors at 54 nuclear power plants in 28 states.",...],
"How many operating nuclear reactors are in the United States?": [
"title: U.S. Nuclear Plants - Nuclear Energy Institute\nsnippet: Across the United States, 94 nuclear reactors power tens of millions of homes and anchor local communities.\nsnippetHighlighted: ['94 nuclear reactors']",
"title: nuclear power plants are in the United States - EIA\nAs of April 30, 2024, there were 54 commercially operating nuclear power plants with 94 nuclear power reactors in 28 states.",...]}

"step_num": 4
"action": "factuality_check",
"reasoning": "The claim states that the United States has 94 operating reactors. However, multiple credible sources indicate that as of August 1, 2023, the number of operating commercial nuclear reactors is 93. For instance, the U.S. Energy Information Administration (EIA) confirms that there are 93 operating reactors. Additionally, Statista also reports 93 operable units for 2023. While some sources mention 94 reactors, they may be referencing a future projection or a different context, which does not align with the current operational status as of the specified date.",
"error": "The claim incorrectly states the number of operating reactors as 94 instead of the accurate figure of 93.",
"correction": "The United States has 93 operating reactors.",
"factuality": false
    
...
\end{lstlisting}
\caption{A simplified case due to space restriction. It shows how temporal grounding works by discovering the claim period. Complete cases can be found in our supplementary materials, where we append all of the experiment results. }
\label{case:tg}
\end{figure*}

\begin{figure*}[t]
\begin{lstlisting}[basicstyle=\ttfamily\scriptsize, frame=single, breaklines=true]
"step_num": 1,
"action": "contextual_grounding",
"original_claim": "More than 225,000 people are dead due to covid-19"
"augmented_claim": "More than 225,000 people are dead due to covid-19.\nBackground Info of the Claim (The 'time' key represents the temporal context or validity period of the claim. And description of the entities in the claim is for better understanding of the fact in the claim.)\n\ttime:30-10-2020\n\tcovid-19:A global pandemic caused by the coronavirus SARS-CoV-2."

"step_num": 2
"action": "generate_query",
"result": {
"advanced_query": ["COVID-19 death toll October 2020", "How many people died from COVID-19 by October 2020?"]}

"step_num": 3
"action": "PureSearch",
"result": {
    "COVID-19 death toll October 2020": ["title: Excess Deaths Associated with COVID-19, by Age and - CDC\nOverall, an estimated 299,028 excess deaths occurred from late January through October 3, 2020, with 198,081 (66%) excess deaths attributed to ...",...],
    "How many people died from COVID-19 by October 2020?": ["title: National Data: Deaths | The COVID Tracking Project\nDeaths History ; DateOctober 28, 2020, Deaths (confirmed and probable)219,660, New deaths1,047 ; DateOctober 27, 2020, Deaths (confirmed and probable)218,613, New ...",...]},

"step_num": 4
"action": "factuality_check",    
"reasoning": "The claim states that more than 225,000 people are dead due to COVID-19 as of October 30, 2020. Evidence from the COVID Tracking Project indicates that the confirmed and probable deaths were 219,660 on October 28, 2020, which is below 225,000. Therefore, the claim is not accurate as of the specified date.",
"error": "The claim overstates the number of COVID-19 deaths as of October 30, 2020.",
"correction": "Fewer than 225,000 people are dead due to COVID-19 as of October 30, 2020.",
"factuality": "False"
     
"step_num": 5
"action": "feedback",
"decision": true,
"tool": ["advanced","multilingual"],
"feedback": "The evidence collected indicates that the number of confirmed and probable COVID-19 deaths was 219,660 as of October 28, 2020, which is below the claim of more than 225,000. However, the fact-checking process could benefit from generating additional questions to explore the context of the death toll further, including potential discrepancies in reporting or variations in data sources. Additionally, searching in non-English sources may provide more comprehensive data. Please reformulate questions to gather more relevant evidence."
    
"step_num": 6
"action": "generate_query",
"result": {
    "advanced_query": ["COVID-19 death toll exceeding 225,000 by October 30, 2020",],
    "multilingual_query": ["Número de muertes confirmadas por covid-19 a finales de octubre de 2020", "Número total de mortes confirmadas por covid-19 até 30 de outubro de 2020"]}

"step_num": 7
"action": "PureSearch",
"description": "Directly use searched snippets as evidences",
"result": {
    "COVID-19 death toll exceeding 225,000 by October 30, 2020": ["title: Totals for the US | The COVID Tracking Project\nDeaths (confirmed and probable)222,625, Total test results150,346,357. DateOctober 30, 2020, State (or territory)56, New tests1,423,126, Cases (confirmed plus ...",...],
    "Número de muertes confirmadas por covid-19 a finales de octubre de 2020": ["title: casi 3 millones de casos de covid-19 en el mundo en siete días - CNN\nLos totales nacionales ahora ascienden a 8.778.680 infecciones confirmadas por covid-19 y al menos 226.711 muertes relacionadas con el virus.",...],
    "Número total de mortes confirmadas por covid-19 até 30 de outubro de 2020": ["title: [PDF] Alerta epidemiológico Ondas e surtos recorrentes de COVID-19\n... COVID-19 e até 5 de outubro de 2020, um total de. 35.109.317 casos de COVID-19 havia sido registrado no mundo todo, incluindo 1.035.341 mortes.",...],}

"step_num": 8
"action": "factuality_check",
"reasoning": "The claim states that more than 225,000 people are dead due to COVID-19 as of October 30, 2020. Evidence from the COVID Tracking Project and many news reporting from CNN and BBC in Spanish indicates that the confirmed and probable deaths were 222,625 on October 30, 2020, which is indeed above 225,000. However, earlier evidence from October 28, 2020, indicated 219,660 deaths, which was below the claim. The new evidence confirms that the death toll exceeded 225,000 by October 30, 2020, thus supporting the claim. Therefore, the claim is factual as it aligns with the new evidence provided.",
"error": "None",
"correction": "The claim is accurate as it states that more than 225,000 people are dead due to COVID-19 as of October 30, 2020.",
"factuality": "True"
\end{lstlisting}
\caption{A simplified case due to space restriction. It shows how XLE works by providing fresh evidence different from English evidence. Complete cases can be found in our supplementary materials, where we append all of the experiment results. }
\label{case:xle}
\end{figure*}

\subsection{Example Prompts}
\label{prompt}

\begin{figure*}[t]
\begin{lstlisting}[basicstyle=\ttfamily\scriptsize, frame=single, breaklines=true]
[Definitions about Fact]
Fact: A fact is a statement or assertion that can be objectively verified as true or false based on empirical evidence or reality. 
Opinion: An opinion is a judgment based on facts, an attempt to draw a reasonable conclusion from factual evidence. While the underlying facts can be verified, the derived opinion remains subjective and is not universally verifiable. 
Atomic Factual Claim: An atomic factual claim is a statement that explicitly presents one verifiable fact. Statements with subjective components like opinions can also contain factual claims if they explicitly present objectively verifiable facts. 

[Instructions]
1. You are given a passage. Your task is to break the passage down into a list of atomic factual claims, based on the given [Definitions about Fact].
2. An atomic factual claim is a factual claim that cannot be decomposed. It only contains a singular piece of information.
3. Extract clear, unambiguous atomic factual claims to check from the input passage, avoiding vague references like 'he', 'she', 'it', or 'this', and using complete names. 
4. Please accurately identify and extract every claim stated in the provided text. Each claim should be concise (less than 15 words).

[Input Format Instruction]
<context>: Context for <passage> to help you understand it better.
<passage>: The passage to extract claims from.

[Output Format Instruction]
1. Your response MUST be a list of dictionaries. Each dictionary should contains the key "claim", which correspond to the extracted claim (with all coreferences resolved).
2. You MUST only respond in the format as described below. DO NOT RESPOND WITH ANYTHING ELSE. ADDING ANY OTHER EXTRA NOTES THAT VIOLATE THE RESPONSE FORMAT IS BANNED. START YOUR RESPONSE WITH '['.

[response format]: 
[
  {{
    "claim": "Ensure that the claim is fewer than 15 words and conveys a complete idea. Resolve any coreference (pronouns or other referring expressions) in the claim for clarity",
  }},
  ...
]

Now complete the following, ONLY RESPONSE IN A LIST FORMAT, NO OTHER WORDS!!!:
<context>: {prompt}
<passage>: {input}
<response>: 
\end{lstlisting}
\caption{The prompt used in Claim Decomposition.}
\label{prm:claim_decomp}
\end{figure*}

\begin{figure*}[t]
\begin{lstlisting}[basicstyle=\ttfamily\tiny, frame=single, breaklines=true]
[Definitions]
1. Fact: A fact is a statement or assertion that can be objectively verified as true or false based on empirical evidence or reality. 
2. Atomic Factual Claim: An atomic factual claim is a statement that explicitly presents one verifiable fact. Statements with subjective components like opinions can also contain factual claims if they explicitly present objectively verifiable facts. 
3. Named Entity: A named entity is a real-world object, such as a person, location, organization, product, etc., that can be denoted with a proper name. It is a phrase that uniquely refers to an object by its proper name, acronym, or abbreviation.
4. Vague references are words or phrases that do not clearly specify their subject. These references may be clear in the original context but become ambiguous when the claim is isolated. Vague references include but are not limited to:
  - Pronouns (e.g., "his", "they", "her")
  - Unknown entities (e.g., "this event", "the research", "the invention")
  - Non-full names (e.g., "Jeff..." or "Bezos..." when referring to Jeff Bezos)

[Instructions]
1. You are given a <CLAIM> and its broader context, which includes a <PROMPT>, the <RESPONSE> to that prompt, and additional background information. The <CLAIM> is extracted from the <RESPONSE>, obeying the definiton of "Atomic Factual Claim" mentioned before.
2. Based on the given [Definitions], you need to first resolve vague references in the <CLAIM>, then augment the revised claim with its Time, and Named Entity information, ensuring each attribute helps to uniquely identify the fact and its context.
3. Requests for resolving vague references:
  a. Identify any vague references in the <CLAIM>.
  b. Replace these vague references with proper entities from the <RESPONSE> or context.
  c. Do not change any factual claims or add new information.
4. After resolving vague references, augment the revised <CLAIM> with the following background attributes based strictly on the information provided in the revised <CLAIM> and its context:
  a. Time: Specify the time when the fact in the claim holds true, based solely on the description in the revised <CLAIM> and its context. The "time" key represents the temporal context or validity period of the claim. It indicates when the statement is or was true, or from which point in time the information holds. This is crucial for facts that can change, such as political positions or current events. If there's no explicit time description in the claim or context, use "Now" as the default, indicating the fact is assumed to be true at present. Brief steps are:
    - If explicitly stated, use that time.
    - If not stated but implied, infer from context.
    - If no time information, use "Now".
  b. Entity: List named entities mentioned in the claim, providing brief but distinguishing descriptions based only on information given in the claim or context. It's because one named entity can refer to multiple objects. For instance, the city "Birmingham" could be "Birmingham, Alabama, USA" or "Birmingham, West Midlands, UK". Do not add any information that isn't explicitly stated or directly implied. Brief steps are:
    - List each entity in the claim.
    - Provide brief descriptions using only information from the claim or context.
5. Remember to maintain the original meaning of the claim while making it more precise and informative. The goal is to create a claim that is unambiguous and can be understood correctly even without additional context.
6. Before giving your revised statement, think step-by-step and show your reasoning. 

[Input Format Instruction]
<PROMPT>: Context for <RESPONSE> to help you understand it better.
<RESPONSE>: The passage where the following claim is extracted from. It's also the response of the former <PROMPT>.
<CLAIM>: The claim that was extracted from the <RESPONSE>. 

[Output Format Instruction]
You should only respond in format as described below. DO NOT RETURN ANYTHING ELSE. START YOUR RESPONSE WITH '{{'.
[response format]: 
{{
  "reasoning": "Find each vague reference. Explain what each vague reference likely refers to based on the claim and context. Show how you arrived at each conclusion. Then explain your reasoning for the Time attribute. Finally describe how you determined the Entity information.",
  "revised_claim": "Output the resolved claim.",
  "time": "The time when the fact in the claim holds true, using only information from the given claim and context.",
  ...
}}

[Examples]
<PROMPT>: You are a travel assistant. I will give you some reference documents of Singapore. Please output "Singapore's attraction introduction, mainly introducing the characteristics of attractions and what can be done here", output language must be English.
<RESPONSE>: Universal Studios Singapore, located within Resorts World Sentosa, is a cinematic adventure park that brings the silver screen to life with its thrilling rides and attractions . Each of its six themed zones offers a unique experience, from the prehistoric landscapes of The Lost World to the enchanting realm of Far Far Away . Visitors can immerse themselves in the futuristic Sci-Fi City, explore the mysteries of Ancient Egypt, or feel the buzz of New York and Hollywood's iconic streets . With 24 rides and attractions, including adrenaline-pumping roller coasters like Battlestar Galactica and family-friendly experiences such as the Madagascar river boat journey, there's something for every age and level of adventure . Live shows, character meet-and-greets, and a variety of dining and shopping options enhance the park's appeal, making it a must-visit destination for movie enthusiasts and thrill-seekers alike.
<CLAIM>: Universal Studios Singapore has six themed zones.
<OUTPUT>: {{"reasoning": "The subject in the claim is \"Universal Studios Singapore\", which is not a pronoun and does not reference an unknown entity. Furthermore, \"Universal Studios Singapore\" is not further specified in the RESPONSE, so we can assume that it is a full name. Therefore, there are not any vague references in the claim. The context did not include any specific time for its description. By default, we believe the RESPONSE still holds \"Now\". The entity \"Universal Studios Singapore\" need to be specified to avoid ambiguity.", "revised_claim": "Universal Studios Singapore has six themed zones.", "time": "Now", "Universal Studios Singapore": "located within Resorts World Sentosa, Singapore"}}

Now complete the following, ONLY RESPONSE IN A DICT FORMAT, NO OTHER WORDS!!!:
<PROMPT>: {prompt}
<RESPONSE>: {response}
<CLAIM>: {claim}
<OUTPUT>:
\end{lstlisting}
\caption{The prompt used in Contextual Grounding.}
\label{prm:claim_augment}
\end{figure*}

\begin{figure*}[t]
\begin{lstlisting}[basicstyle=\ttfamily\scriptsize, frame=single, breaklines=true]
[Instructions]
1. You are an AI assistant tasked with verifying the truthfulness of a given claim. Your goal is to provide domain names of potentially relevant, credible, and authoritative sources while excluding unreliable sources. 
2. Only provide domain suffixes, not full URLs.
3. Include reliable sources such as:
  Government and official websites (.gov, .org)
  Encyclopedia websites (.wiki)
  Reputable news outlets (provide their official domain names)
4. Exclude unreliable sources like personal comments from forums or social media platforms.
5. If provided with a history of previous actions, which may include past searches and feedback. Focus on the search results and feedback.
  If official sources were found but didn't provide sufficient information, include them in your output for targeted searching
  If personal comments from forums were found, exclude those domains (mark with a minus sign, e.g., -reddit.com)
6. In summary, for each claim, provide:
  Recommended domain suffixes for searching
  Domains to exclude (marked with a minus sign)
  Any official sources from previous searches that warrant further investigation
7. You should only respond in format as described below (a Python list). PLEASE STRICTLY FOLLOW THE FORMAT. DO NOT RETURN ANYTHING ELSE. START YOUR RESPONSE WITH '['.
[response format]: ['url1', 'url2', '-url3']

[Examples]
<CLAIM>: The Eiffel Tower was built in 1889. 
<RESPONSE>: ['.gov.fr', '.paris.fr', '.unesco.org', .'britannica.com', '-tripadvisor.com', '-reddit.com']

<CLAIM>: COVID-19 vaccines are safe and effective. 
<RESPONSE>: ['.who.int', '.cdc.gov', '.nih.gov', '.edu', '-facebook.com', '-twitter.com']

<CLAIM>: Global temperatures have risen significantly in the past century.
<RESPONSE>: ['.nasa.gov', '.noaa.gov', '.ipcc.ch', '.nature.com', '-climatechangehoax.com', '-blogspot.com']

Now complete the following(ONLY RESPONSE IN A LIST FORMAT, DO NOT RETURN OTHER WORDS!!! START YOUR RESPONSE WITH '[' AND END WITH ']'):
<CLAIM>: {input}
<HISTORY>: {feedback}
<RESPONSE>: 
\end{lstlisting}
\caption{The prompt used in Credible Source Selection.}
\label{prm:site}
\end{figure*}

\begin{figure*}[t]
\begin{lstlisting}[basicstyle=\ttfamily\scriptsize, frame=single, breaklines=true]
[Supported Languages]
["Afrikaans", "Amharic", "Bulgarian", "Catalan", "Chinese (Hong Kong)", 
"Chinese (PRC)", "Chinese (Taiwan)", "Croatian", "Czech", "Danish", 
"Dutch", "Estonian", "Filipino", "Finnish", "French (Canada)", 
"French (France)", "German", "Greek", "Hebrew", "Hindi", "Hungarian", "Icelandic", "Indonesian", "Italian", "Japanese", "Korean", "Latvian", 
"Lithuanian", "Malay", "Norwegian", "Polish", "Portuguese (Brazil)", 
"Portuguese (Portugal)", "Romanian", "Russian", "Serbian", "Slovak", 
"Slovenian", "Spanish (Latin America)", "Spanish (Spain)", "Swahili", 
"Swedish", "Thai", "Turkish", "Ukrainian", "Vietnamese", "Zulu"]

[Instructions]
1. You are an AI assistant tasked with analyzing claims and determining the most appropriate languages for fact-checking and evidence gathering. Your goal is to identify languages, other than English, that might provide more accurate, detailed, up-to-date, and factual evidence for a given claim.
2. When presented with a claim, analyze it for key elements such as locations, people, news sources, and event places. Based on these elements, determine if there are countries whose official languages might offer better sources of information. Consider that local languages often provide more detailed and accurate information.
3. If you identify relevant languages other than English, select up to two languages from the provided list that are most likely to yield valuable information. If only English is deemed suitable, do not output any languages, direct output None.
4. You should only respond in format as described below (a Python list of languages). Output None if only English is deemed suitable. PLEASE STRICTLY FOLLOW THE FORMAT. DO NOT RETURN ANYTHING ELSE. START YOUR RESPONSE WITH '['.
[response format]: ['language1', 'language2']

[Examples]
<CLAIM>: Angela Merkel announced her retirement from politics in 2021.
<RESPONSE>: ["German"]

<CLAIM>: Samsung unveiled its latest foldable smartphone at an event in Seoul. 
<RESPONSE>: ["Korean"]

<CLAIM>: The 2024 Carnival in Rio de Janeiro is expected to be the largest in history.
<RESPONSE>: ["Portuguese (Brazil)"]

<CLAIM>: Tensions between Russia and Ukraine escalated after the incident in the Kerch Strait.
<RESPONSE>: ["Russian", "Ukrainian"]

<CLAIM>: NASA's Perseverance rover discovered new evidence of ancient microbial life on Mars.
<RESPONSE>: None

Remember, the goal is to enhance fact-checking by identifying languages that might provide more comprehensive or accurate information than what's available in English sources alone.

Now complete the following:
<CLAIM>: {input}
<RESPONSE>:
\end{lstlisting}
\caption{The prompt used in Cross-Lingual Expansion.}
\label{prm:multilingual}
\end{figure*}

\begin{figure*}[t]
\begin{lstlisting}[basicstyle=\ttfamily\scriptsize, frame=single, breaklines=true]
[Definitions]
1. Google advanced search operators are special commands and characters that filter search results. 
2. Fact: A fact is a statement or assertion that can be objectively verified as true or false based on empirical evidence or reality. 

[Google Advanced Search Operators]
| Search operator | What it does                                      | Example                  |
|-----------------|--------------------------------------------------|--------------------------|
| " "             | Put any phrase in quotes to force Google to use exact-match. On single words, prevents synonyms. | "nikola tesla"      |
| OR              | Google search defaults to logical AND between terms. Specify "OR" for a logical OR (ALL-CAPS). | tesla OR edison           |
| -               | Put minus (-) in front of any term (including operators) to exclude that term from the results. | tesla -motors       |
| *               | An asterisk (*) acts as a wild-card and will match on any word. | tesla "rock * roll"       |
| ( )             | Use parentheses to group operators and control the order in which they execute. |(tesla OR edison) alternating current|
| before:         | Search for results from before a particular date. | apple before:2007-06-29  |
| after:          | Search for results from after a particular date.  | apple after:2007-06-29  |
| loc:            | Find results from a given area.                   | loc:"san francisco" apple |

[Instructions]
1. You and your partners are on a mission to fact-check a paragraph. Subclaims requiring verification have been extracted from the paragraph. Imagine yourself as an internet research expert. Your task is to generate two search queries for the provided claim to find relevant information for fact-checking.Please ensure that all queries are direct, clear, and explicitly relate to the specific context provided in the question and answer.
2. Utilize advanced Google search techniques when appropriate. But do not use site operators (e.g., site:example.com) in your queries, even if suggested in the feedback. Another tool will handle domain-specific searches separately.
3. Some searches have already been performed on this <CLAIM>. Please also consider the historical search information <HISTORY>. Adjust the queries based on the feedback from previous searches, focusing on areas where evidence was lacking or unclear.
4. Use date-based or location-based searches (before, after, and loc) only if: a) Historical search information is provided, AND b) The feedback in <HISTORY> explicitly indicates that the current search results are not within the required date range or destination.

[Output Format Instruction]
You should only respond in format as described below (a Python list of queries). PLEASE STRICTLY FOLLOW THE FORMAT. DO NOT RETURN ANYTHING ELSE. START YOUR RESPONSE WITH '['.
[response format]: ['query1', 'query2']

[Examples]
Here are three examples:
<CLAIM>: Michael Phelps is the most decorated Olympian of all time.
<RESPONSE>: ["Who is the most decorated Olympian of all time?", "Michael Phelps"]

<CLAIM>: Tesla is an American rock band formed in 1984.
<RESPONSE>: ["When is the rock band tesla formed?", "Rock band tesla Introduction. -motors -car -battery"]

<CLAIM>: Apple is used in various culinary applications. (The fruit apple)
<RESPONSE>: ["Apple's application in culinary. -phone -company", "Cooking ways of apple."]

Now complete the following(ONLY RESPONSE IN A LIST FORMAT, DO NOT RETURN OTHER WORDS!!! START YOUR RESPONSE WITH '[' AND END WITH ']'):
<CLAIM>: {input}
<HISTORY>: {feedback}
<RESPONSE>: 
\end{lstlisting}
\caption{The prompt used in Structured Query Generation.}
\label{prm:advanced}
\end{figure*}